\documentclass[10pt,twocolumn,letterpaper]{article}

\usepackage{iccv}
\usepackage{times}
\usepackage{epsfig}
\usepackage{graphicx}
\usepackage{amsmath}
\usepackage{amssymb}
\usepackage{subfig} 
\usepackage{tabulary}
\usepackage{colortbl}
\usepackage{booktabs}

\usepackage{microtype}      
\usepackage[dvipsnames,table,xcdraw]{xcolor}
\usepackage[numbers,sort]{natbib}
\usepackage{multirow}
\usepackage{floatrow}
\usepackage{caption}
\usepackage{pifont}

\iccvfinalcopy 

\definecolor{mygray}{gray}{.92}

\newfloatcommand{capbtabbox}{table}[][\FBwidth]
\newcommand{\cmark}{\ding{51}}%
\newcommand{\xmark}{\text{\ding{55}}}

\newlength\savedwidth
\ificcvfinal\fi
\usepackage[pagebackref=true,breaklinks=true,letterpaper=true,colorlinks,bookmarks=false]{hyperref}

\usepackage{cleveref}

\begin{document}

\title{Masked autoencoders are effective solution to transformer data-hungry}

\author{Jiawei Mao \quad Honggu Zhou  \quad Xuesong Yin {\thanks{Corresponding author.}}  \quad Yuanqi Chang  \quad Binling Nie \quad Rui Xu   \\ 
School of Media and Design, Hangzhou Dianzi University, Hangzhou, China \qquad \\
{\tt\small\{jiaweima0,hongguzhou,yinxs,yuanqichang,binlingnie,211330017\}@hdu.edu.cn }\\
}

\maketitle
\ificcvfinal\thispagestyle{empty}\fi

\begin{abstract}
Vision Transformers (ViTs) outperforms convolutional neural networks (CNNs) in several vision tasks with its global modeling capabilities. 
However, ViT lacks the inductive bias inherent to convolution making it require a large amount of data for training. 
This results in ViT under performing CNNs on small datasets like medicine and science. 
We experimentally found that masked autoencoders (MAE) can make the transformer focus more on the image itself, thus alleviating the data-hungry issue of ViT to some extent. 
Yet the current MAE model is too complex resulting in over-fitting problems on small datasets. 
This leads to a gap between MAEs trained on small datasets and advanced CNNs models still. 
To this end, we investigate how to reduce the decoder complexity in MAE and find a more suitable architectural configuration for it with small datasets. 
Besides, we additionally design a location prediction task and a contrastive learning task to introduce localization and invariance characteristics for MAE. 
Our contrastive learning task not only enables the model to learn high-level visual information, but also allows the training of MAE's class token, 
which is not considered in most MAE improvement efforts. 
Extensive experiments on standard small datasets and medical datasets with few samples show that our method can achieve the state-of-the-art performance  
compared with the current popular masked image modeling (MIM) and vision transformers for small datasets.
The code and models are available at ~\url{https://github.com/Talented-Q/SDMAE}.
\end{abstract}

\section{Introduction}

CNNs were once dominant in computer vision by the characteristics of localization, translation invariance and hierarchy of convolution. 
Since Dosovitskiy et al.\cite{dosovitskiy2020image} introduced transformer from the field of natural language processing (NLP) to computer vision domain, a series of visual transformer works\cite{touvron2021training,zhou2021deepvit,yuan2021tokens,chen2021crossvit,heo2021rethinking,graham2021levit} represented by ViT have been developed rapidly in recent years. 
ViTs have emerged as an alternative to convolution in several vision tasks with its powerful global modeling capabilities.

\begin{figure}[t]
   \centering
   \includegraphics[width=1\linewidth]{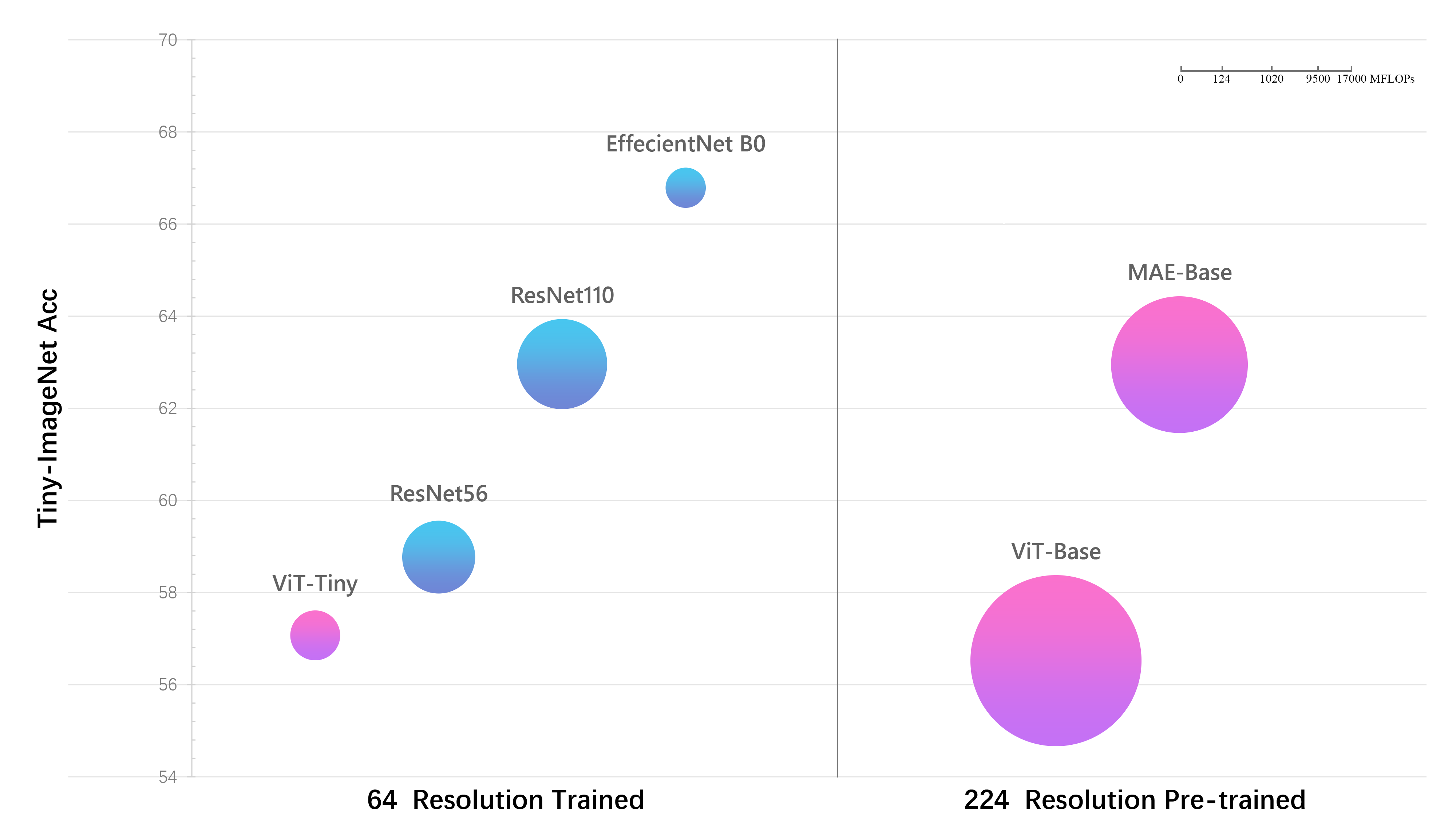}
    \caption{Comparison results of ViTs and CNNs on Tiny-ImageNet. We can observe that ViT-Base and ViT-Tiny do not perform as well as CNNs, both at small and large resolutions. 
    MAE pre-training successfully improves ViT-Base's accuracy on Tiny-ImageNet, making it surpass ResNet56 and reach the performance of ResNet110. 
    However, the MAE still has a certain 	gap 	compared with EffecientNet B0.}
    \label{fig1}
 \end{figure}

Although ViTs have shown such excellent performance in the field of computer vision, the lack of induction bias of CNNs in transformers causes them to need to learn these properties from large-scale datasets. 
For example, ViT and its variants are typically trained using large-scale datasets such as ImageNet-1k/22k \cite{deng2009imagenet} or JFT-300M \cite{sun2017revisiting}, which contains 303 million images. 
However, many subjects, such as medicine and science, are not equipped with large-scale datasets due to privacy protection, limited exploration equipment and other factors. 
This makes them fail to train a transformer network with good analysis capability. Therefore, methods that can train a good transformer on small datasets are particularly important.

To make ViTs perform better on small datasets, Lee et al.\cite{lee2021vision} proposed Shifted Patch Tokenization (SPT) and Locality Self-Attention (LSA) to enhance the local induction bias of ViT. 
Hassani et al.\cite{hassani2021escaping} designed a more compact ViT architecture to suppress the issue of model over-fitting on small datasets. 
Gani et al.\cite{gani2022train} introduce local, invariant, and hierarchical properties to ViT by designing self-supervised tasks. 
Despite the fact that all these approaches improve ViT performance on small datasets to some extent, they all used smaller ViT configuration models. 
This prevents them from achieving uniformity in ViT training on large and small datasets. 
We prefer to find a training method that does not require changes to the ViT configuration in small dataset.

Recently MAE\cite{he2022masked} has effectively improved the performance of ViT downstream tasks by recovering the mask tokens pixel information. 
While MAE has been applied to many visual tasks, to our knowledge, no work has been done to investigate whether MAE can solve the data-hungry issue of transformer. 
We argue that MAE can somewhat reduce the dependence of ViT models with standard configuration on large-scale datasets by focusing on the images themselves. 
To verify this conjecture, we train the standard MAE on a representative small dataset Tiny-ImageNet\cite{tavanaei2020embedded}. 
Figure~\ref{fig1} demonstrates that MAE effectively improves the performance of ViT on small datasets, but still falls short of the advanced CNN. 
This is due to the fact that the standard configuration of the MAE decoder is too complex for small datasets suppressing the performance of the ViT encoder. 
Therefore, we conduct a lot of decoder weakening experiments to find an architecture configuration for MAE's decoder that is more suitable for small datasets. 
In addition, we design a location prediction task as well as a contrastive learning task for MAE to introduce the localization and invariance features of CNNs. 
Our location prediction task consists only of a tiny location predictor with negligible parameters and without any human assisted labeling. 
Our contrastive learning task also allows MAE to learn advanced semantic information about the images and to train class token specifically for MAE. 
This is because MAE and its refinement efforts often do not have a training task specifically for class token in the pre-training design. 
However, class token is indispensable in the fine-tuning phase, so a pre-training task for class token is necessary. 
We combine this masked image modeling with the above designs and call it Small Dataset Masked Autoencoders (SDMAE).

Later experimental results show that our SDMAE exhibits state-of-the-art performance on standard small datasets Tiny-ImageNet\cite{tavanaei2020embedded}, CIFAR-100\cite{krizhevsky2009learning}, 
CIFAR-10\cite{krizhevsky2009learning} and SVHN\cite{liao2015competitive} in comparison to vision transformers for small datasets and advanced CNNs. 
Moreover, SDMAE is superior to the popular MIM on these datasets. This indicates that SDMAE is more suitable for standard configuration of ViT models trained on small datasets. 
Besides, SDMAE excels in medical diagnostic tasks with small datasets. This further demonstrates the effectiveness of SDMAE in the practical application on small datasets.

To sum up, our method has the following contributions:
\begin{itemize}
   \item To the best of our knowledge, our SDMAE is the first work to study MAE training ViT on a small dataset. We found the optimal decoder configuration for MAE through extensive experiments to solve the over-fitting problem of MAE decoder on small datasets. Moreover, since our SDMAE does not change the ViT configuration, it achieves unification of ViT training on both large-scale and small datasets.
   \item We design a tiny predictor with negligible parameters to perform the location prediction task of unmasked tokens based on the properties of MAE random masks. This task introduces the localization feature of CNNs for ViT.
   \item Our contrastive learning task design in SDMAE not only introduces CNNs invariant features to ViT, but also enables MAE to train class token and learn high-level semantic information of images. Training of class token is not considered in the MAE and most of its improvement efforts.
\end{itemize}

\section{Related Work}
\subsection{Convolutional Neural Networks}
CNNs play a leading role in the field of computer vision depending on their inductive bias characteristics. 
LeCun first established the rudiment of convolutional neural network training in \cite{lecun1998gradient}. 
The proposal of AlexNet\cite{krizhevsky2017imagenet} further promoted the development of CNNs. Szegedy et al. proposed that GoogleNet\cite{szegedy2015going} greatly improved the utilization of computer resources by increasing the depth of CNN. 
Simonyan et al.\cite{simonyan2014very} studied the depth of CNNs to reduce the parameters of convolution model. 
He et al.\cite{he2016deep} solved the vanishing gradient problem of CNNs by residual linking, thus designing a more compact ResNet. 
Huang et al.\cite{huang2017densely} designed an efficient DenseNet by implementing feature reuse through dense connection. 
EfficientNet\cite{tan2019efficientnet} uses composite coefficients to uniformly scale all dimensions of the model, greatly improving the accuracy and efficiency of CNNs. 
Liu et al.\cite{liu2022convnet} reviewed the CNNs design space from the perspective of transformer through a large amount of experiments and proposed ConvNext with higher accuracy. 
In recent years, CNNs are still the preferred choice for researchers to study small datasets through the characteristics of locality, translation invariance and hierarchy.

\begin{figure*}[t]
   \centering
   \includegraphics[width=1\linewidth]{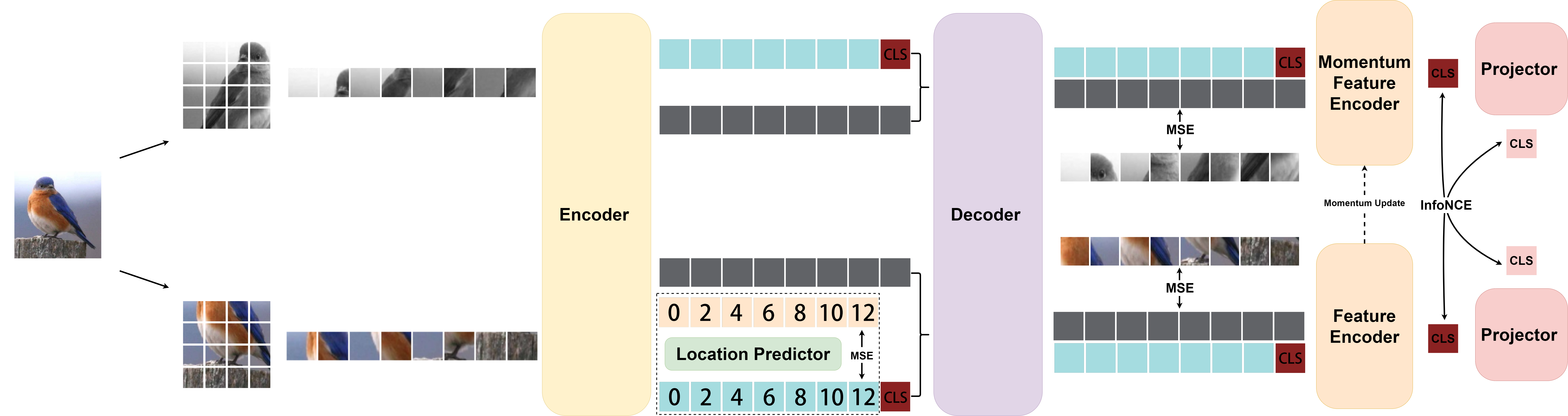}
    \caption{SDMAE detailed process. SDMAE first performs strong data augmentation and weak data augmentation on the input images separately and performs image prediction tasks on both images in the manner of MAE. 
    SDMAE's encoder also needs to perform location prediction task. 
    Lastly, the feature encoder and projection layer of SDMAE perform contrastive task on the class token output by the decoder.}
    \label{fig2}
 \end{figure*}

\subsection{Vision Transformers for small datasets}
Vision Transformers outstanding in many mainstream visual tasks, such as image classification\cite{liu2021swin,wu2021cvt,chu2021twins,chen2021regionvit,wang2021crossformer,wang2021pyramid,yang2022scalablevit,li2022sepvit,tu2022maxvit}, 
semantic segmentation\cite{zhou2022rethinking,ding2022decoupling,wang2022semi,kim2022restr,zhang2022bending,li2022deep,zhang2022semantic,guo2022simt}, 
and object detection\cite{zand2022objectbox,jin2022you,yang2022focal,zhao2022semantic,li2022oriented,zhou2022multi}, depending on its global modeling characteristics. 
Matsoukas et al.\cite{matsoukas2021time} proved that ViTs can also replace CNNs in the medical field. 
However, due to the lack of inductive bias of CNNs, ViTs needs a lot of data for training. 
To solve this issue, Lee et al.\cite{lee2021vision} proposed Shifted Patch Tokenization (SPT) and Locality Self-Attention (LSA) for ViTs. 
Hassani et al.\cite{hassani2021escaping} analyzed the structure of ViTs and improved the representation learning ability of ViTs on small datasets by modifying the ViTs architecture. 
Liu et al.\cite{liu2021efficient} designed a relative position prediction task to regularize ViTs training on small datasets. 
Gani et al.\cite{gani2022train} introduced the inductive bias attribute of CNNs for ViTs through self-supervised task design. 
The research of El Nouby et al.\cite{el2021large} indicates that large-scale datasets are not necessary for the self-supervised pre-training task of the standard configuration ViT model. 
Therefore, we want to find a self-supervised method that can train standard configuration ViT models well on small datasets in order to achieve uniform ViT training on large and small datasets.

\subsection{Masked Image Modeling}
MIM enables the transformer to better learn image representation by recovering the masked tokens. Bao et al. introduced BERT\cite{devlin2018bert} in NLP into the field of computer vision and designed BEiT\cite{bao2021beit}. 
Compared with the previous ViTs pre training methods, this model has achieved better results. 
He et al. proposed MAE\cite{he2022masked}, which greatly improves the model efficiency while enhancing the performance of ViT. 
Xie et al. considered the structure design of MIM and proposed SimMIM\cite{xie2022simmim} with only one layer of MLP decoder. 
Chen et al. proposed SdAE\cite{chen2022sdae} by combining self-distillation with MIM. Dong et al. designed BootMAE\cite{dong2022bootstrapped} to force ViTs to fully learn high-level and low-level visual information in the pre-training stage. 
Chen et al.\cite{chen2022context} designed alignment tasks for the encoder to enhance its representation learning ability. 
Huang et al.\cite{huang2022contrastive} introduced contrastive learning to enhance the discriminability of MAE. Xiao et al.\cite{xiao2022delving} also used MAE in multi label thorax disease classification research. 
Although the improvement of MAE has shown excellent performance, we think that there are still two issues that have been ignored. 
First of all, in order to improve MAE, some works has compromised the advantage of MAE's low computation. 
Second, the class token, which is crucial in MAE fine-tuning, is ignored in the design of pre-training tasks.

\section{Method}

In this section we first review the specific process of MAE in \textcolor{red}{3.1}. Then we experimentally choose the best configuration for MAE's decoder on small datasets in \textcolor{red}{3.2}. 
Finally in \textcolor{red}{3.3} and \textcolor{red}{3.4} we introduce the SDMAE location prediction task and the contrastive learning task, respectively. The overview of SDMAE is shown in Figure~\ref{fig2}.

\subsection{Preliminaries}
\begin{figure}[ht]
   \centering
   \includegraphics[width=1\linewidth]{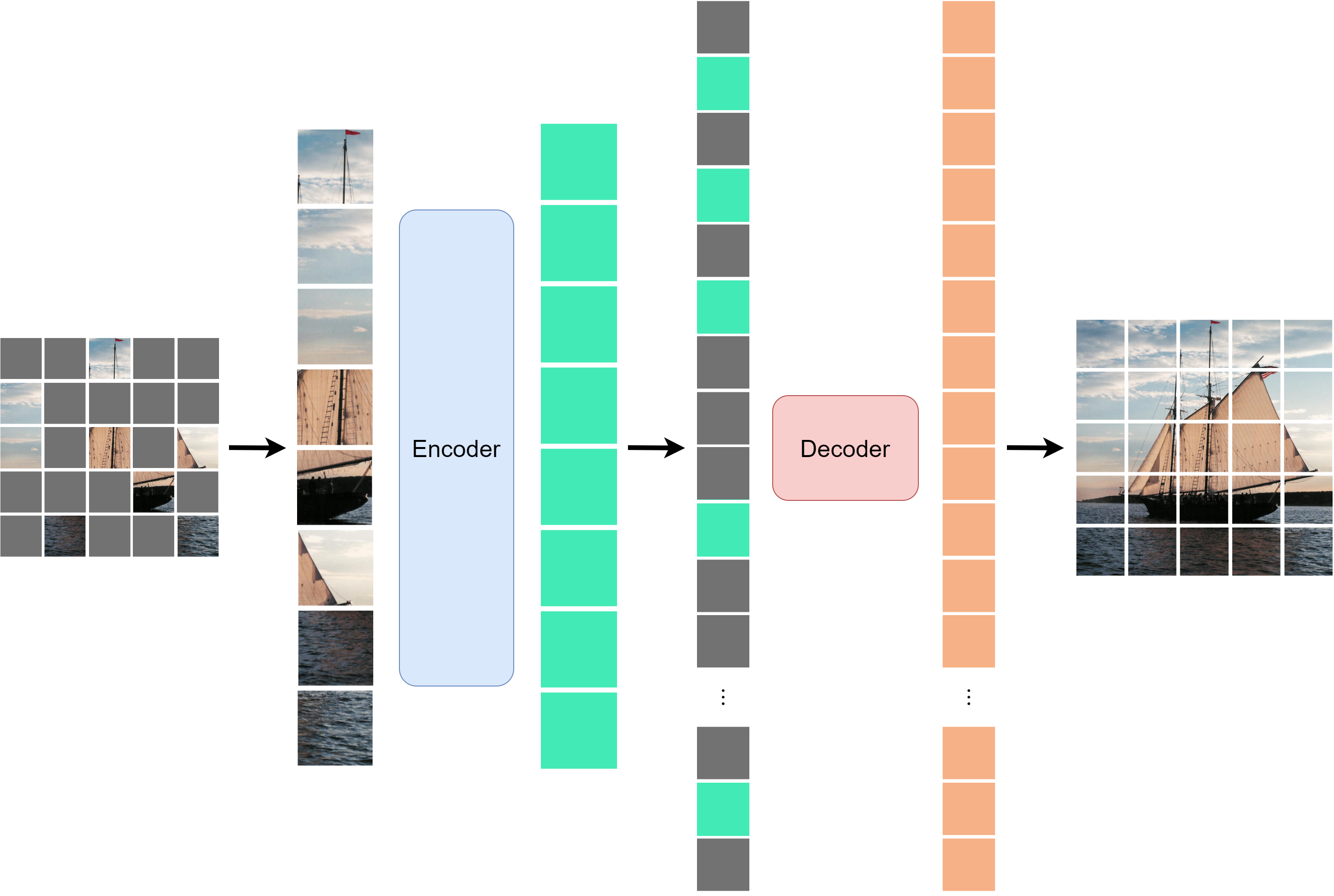}
    \caption{MAE specific workflow. MAE first transforms the image into embedded tokens and then masks these tokens. Visible tokens are processed with an encoder composed of ViT. 
    The encoded tokens are fed to the decoder together with the learnable mask tokens for prediction of the original mask tokens.}
    \label{fig3}
 \end{figure}
 Figure~\ref{fig3} illustrates the specific flow of MAE. MAE is mainly composed of two core designs, the asymmetric encoder-decoder architecture and the high proportion of random masks. 
 Given an input image ${x} \in {\mathbb{R}^{H \times W \times 3}}$, it is first divided into N patches by a $P \times P$  convolution with stride P, and $N = HW/{P^2}$ Next all patches are flattened to one-dimensional sequential tokens ${x_p} \in {\mathbb{R}^{N \times D}}$,
 where $D = {P^2}C$. Then MAE masks these tokens randomly according to the masking ratio $m$. At this moment we obtain the visible token set ${{x}_{vis}} \subseteq {\mathbb{R}^{{N_v} \times D}}$
 and the masked token set ${{x}_{m}} \subseteq {\mathbb{R}^{{N_{m}} \times D}}$, where ${N_v} = (1 - {m}) \times N,{N_{m}} = {m} \times N$.

 The MAE encoder then linearly maps the visible tokens and adds holistic positional embeddings $E_{vis}$ for them. After a series of transformer blocks, we get the encoding ${{z}_{vis}} \subseteq {\mathbb{R}^{{N_v} \times D'}}$ of the visible tokens. 
 Then MAE uses masked tokens ${{z}_{m}}$ composed of learnable vectors to learn the low-level visual information of the image.
 ${{z}_{m}}$ are merged with ${{z}_{vis}}$ in the correct order to get full token set ${{z}_{all}}$ The MAE decoder also adds positional embeddings ${E_{{pos}}}$ to ${z_{all}}$ and feeds them to several transformer blocks to get the final output $y$. 
 MAE eventually selects the tokens $y_m$ of the masked region from $y$ to predict the original set of masked tokens $Y_m$ after normalization. The decoder maps ${{y}_m}$ to ${y'}_m$ through the multi-layer perception (MLP) thus achieving dimension alignment with $y$. 
 Overall the final prediction process can be expressed as:
 \begin{equation}
   \begin{split}
         &{Y_m} = Norm({{{x}}_{{m}}}),\\
         &{L_{{{mse}}}} = MSE(y'_m,{Y_m}),
 \end{split}
   \label{eq1}
 \end{equation}
 where $Norm\left(  \cdot  \right)$ denotes normalization and $MSE\left(  \cdot  \right)$ is the mean squared error\cite{bauer1999empirical}.

 \subsection{Decoder Weakening}

 \begin{figure}[ht]
	\centering
	\includegraphics[width=1\columnwidth]{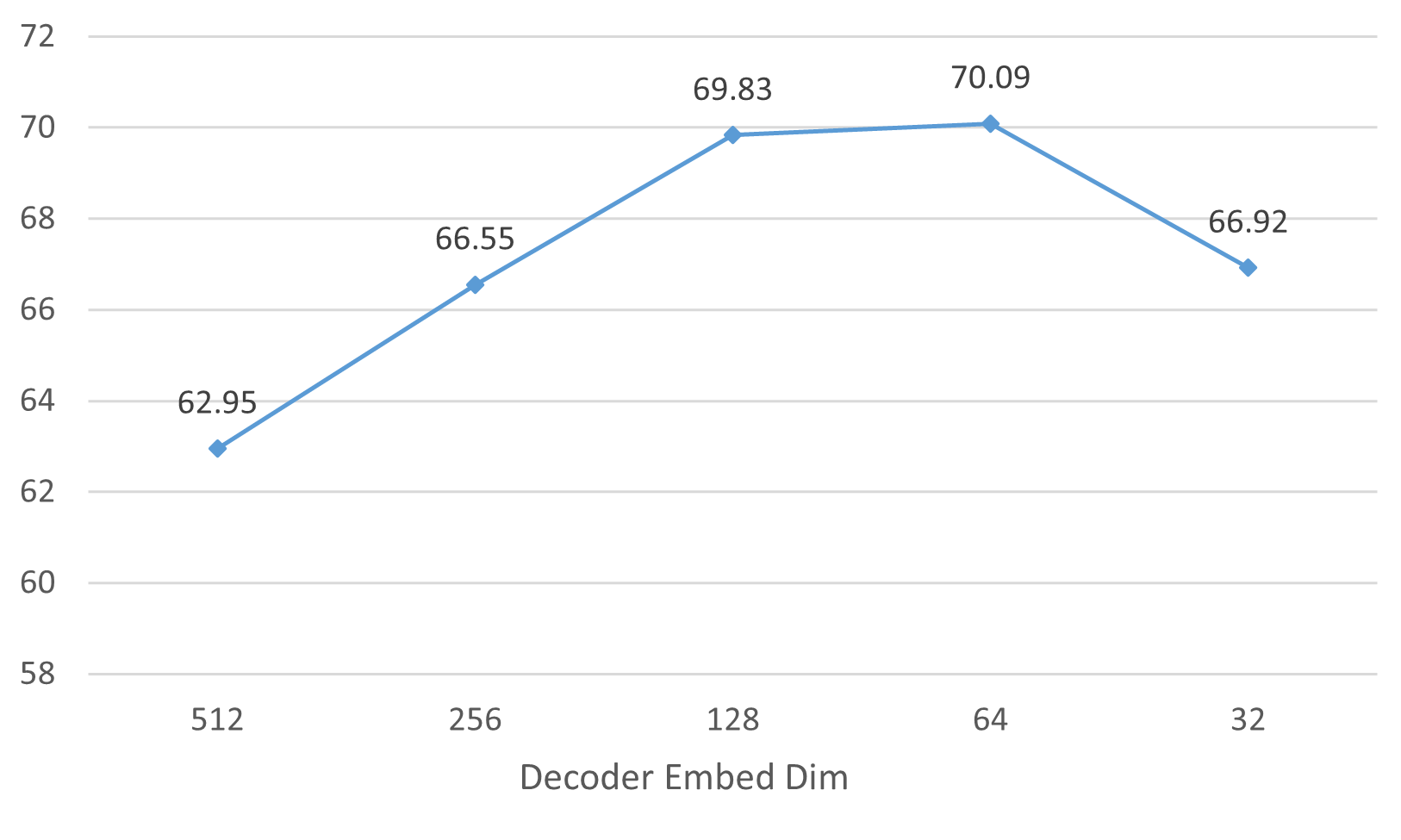}\hspace{5pt}\\
	\includegraphics[width=1\columnwidth]{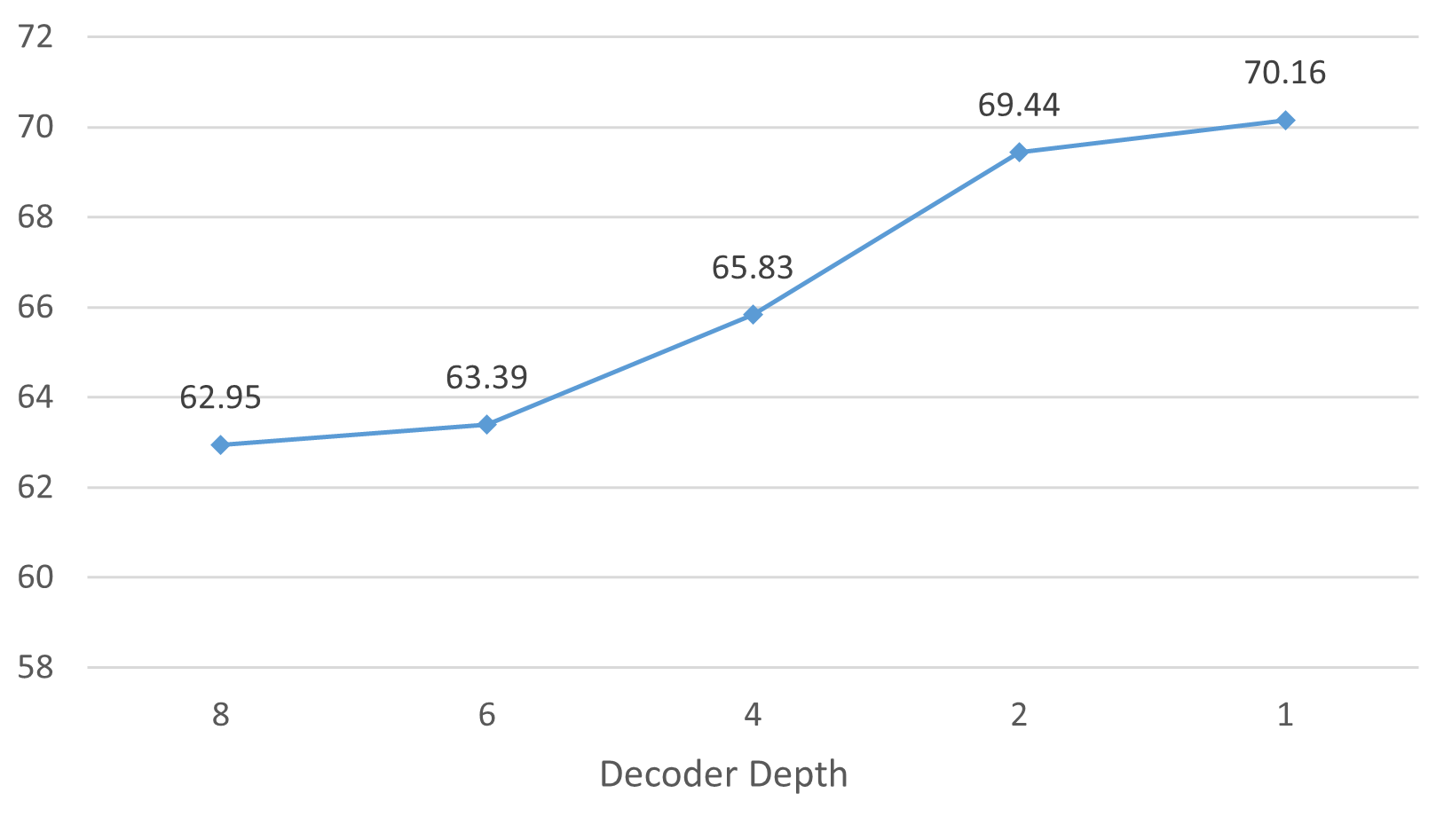}
	\caption{ The effect of decoder depth and embedding dimension on the training of MAE on Tiny-ImageNet. 
   We can find that the MAE performance keeps improving as the decoder keeps weakening. 
   However, MAE performance degrades when the decoder is so weakened that it cannot guarantee the predicted image quality.}
   \label{fig4}
\end{figure}

We found through our experiments that MAE trained on small datasets, while largely reducing ViT's dependence on data volume, still fall short of the current state-of-the-art CNNs. 
We speculate that this is due to the over-complication of the standard configuration MAE decoder, which causes an over-fitting problem that ultimately inhibits the performance of the ViT encoder.

To solve this problem, we weaken the decoder in both dimensionality and depth directions respectively to find the most suitable decoder configuration for MAE training on small datasets. 
Figure~\ref{fig4} show the influence of decoder dimension and depth on MAE training on small datasets, respectively. 
We can find that the performance of MAE on small datasets keeps improving when the decoder dimension and depth are reduced. 
This validates our conjecture that the standard configuration MAE decoder over-fits on small datasets.

\begin{figure}[ht]
   \centering
   \includegraphics[width=1\linewidth]{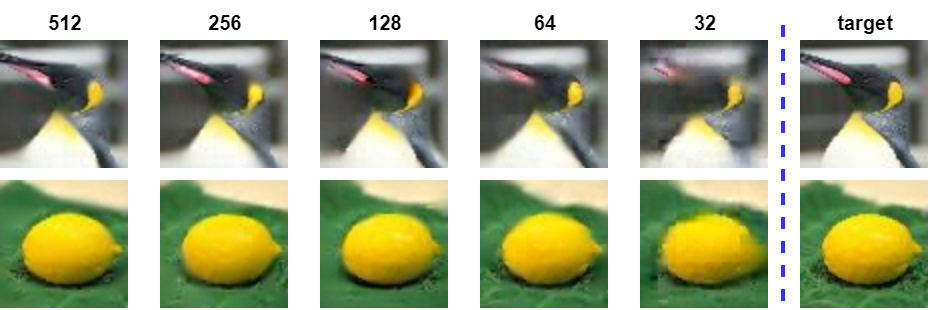}
    \caption{The predicted image in each decoder dimension. 
    We can find that in the case of very low decoder dimension, the quality of the predicted image suffers a serious decline and is almost unrecognizable.}
    \label{fig5}
 \end{figure}

 However, we continue to weaken the decoder and find that the performance of MAE starts to deteriorate instead. 
 We consider that this is caused by the fact that the decoder complexity of MAE is no longer sufficient to recover the image at this time. 
 We can see from Figure~\ref{fig5} that the predicted image quality becomes very poor when the MAE decoder is extremely weakened. 
 This suggests that we cannot continuously weaken the decoder. 
 Therefore, our SDMAE eventually selects a decoder configuration with a depth of 1 and an embedding dimension of 128. We supplement it in section \textcolor{red}{4.3}.

 \subsection{Location Prediction Task}

 The goal of our location prediction task is to introduce spatially localized information for ViT without using additional manual annotations. 
 It is due to the nature of the MIM random mask that our SDMAE can easily achieve this task. 
 Specifically, MAE divides the image sequential tokens into visible token set ${{x}_{vis}} \subseteq {\mathbb{R}^{{N_v} \times D}}$ and masked token set ${{{x}}_{{m}}} \subseteq {\mathbb{R}^{{N_{{m}}} \times D}}$ by random masks with mask ratio $m$. 
 Affected by this operation, the position of each token in the visible token set no longer matches the normal position in the image. 
 Thus, we can predict the location of the encoded visible token set ${{z}_{vis}} \subseteq {\mathbb{R}^{{N_v} \times D'}}$ in the sequential tokens of the image and thus make the encoder learn the spatial local information of the image.

 \begin{figure}[ht]
   \centering
   \includegraphics[width=1\linewidth]{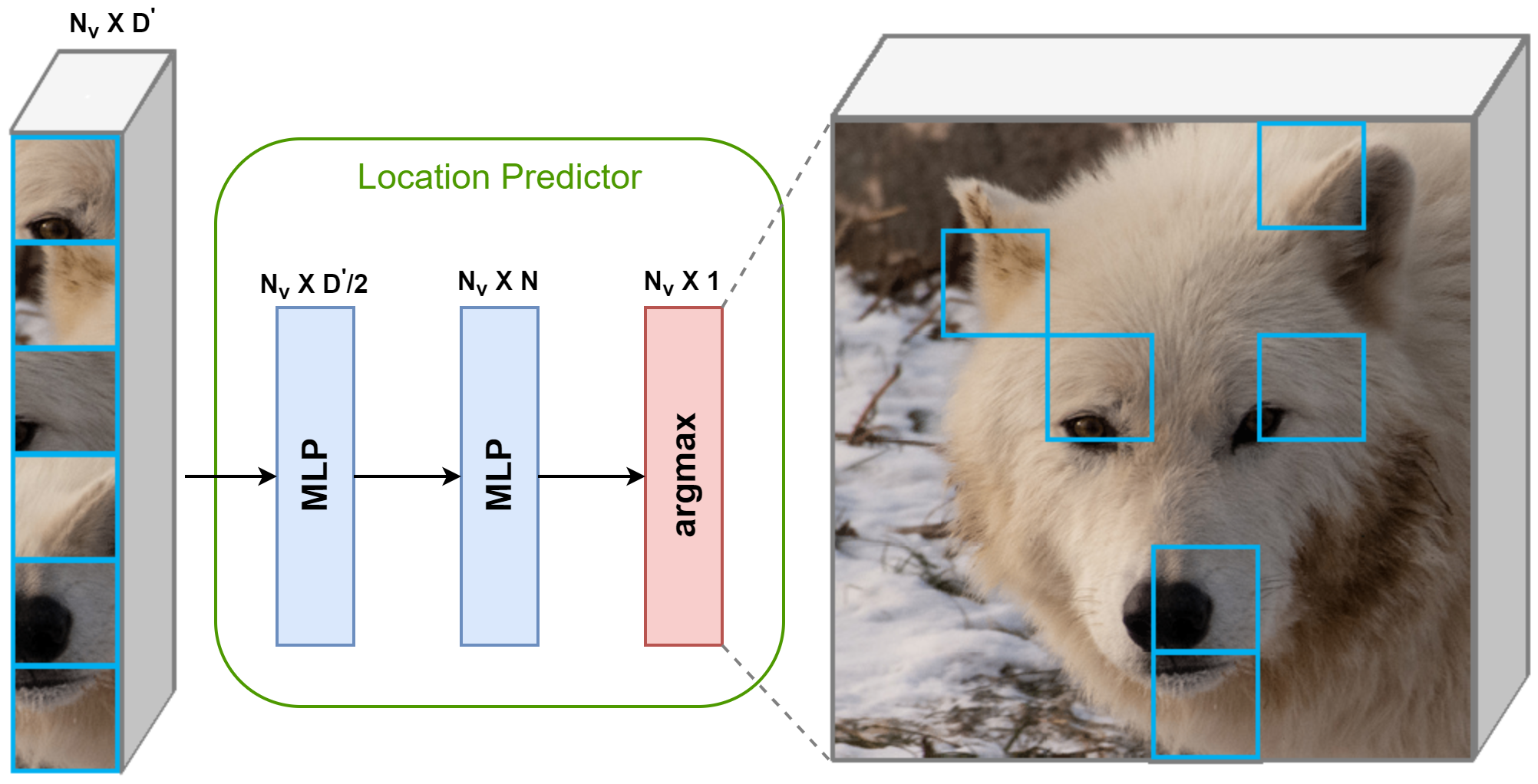}
    \caption{The specific process and model architecture of the location predictor.}
    \label{fig6}
 \end{figure}

 We implement the above task with a tiny predictor with negligible parameters. Our location predictor consists of only two MLP in a squeezed form. 
 Figure~\ref{fig6} shows the specific details of the location predictor. 
 We assume that the position index of each token of the visible token set in the image sequential tokens is ${{t}_i},i \subseteq [1,{N_v}]$ ,
 and the location predictor has prediction result ${{{d}}_i},i \subseteq [1,{N_v}]$ for each token in ${{z}_{vis}}$ (Note that the ${{z}_{vis}}$ here does not contain the class token).  
 However, there is a large numerical difference between $d_i$ and $t_i$, so we cannot simply regress them by mean squared error. 
 Therefore, the details of our approach are to map $d_i$ to an $N_v$-dimensional vector and then regress the maximum index in the $N_v$-dimensional vector against $t_i$ . 
 And the overall location prediction loss function can be further formulated as:
 \begin{equation}
   \begin{split}
         &{{d}} = ({{{z}}_{vis}}{W_a}){W_b},{W_a} \subseteq {\mathbb{R}^{D' \times \frac{{D'}}{2}}},{W_b} \subseteq {\mathbb{R}^{\frac{{D'}}{2} \times {N_v}}},\\
         &{L_{loc}} = MSE(\arg \max (d),t),
 \end{split}
   \label{eq2}
 \end{equation}
 where $\arg \max (.)$ represents the operation that returns the index of the last dimensional maximum.

 \subsection{Contrastive Task}

 To introduce the feature of invariance of CNNs, we design the contrastive task for MAE. 
 Most current MAE combined with contrastive learning efforts\cite{chen2022context,yi2022masked} tend to have the following two issues. 
 First, they encode the full image in addition to the set of visible tokens. This inevitably increases the amount of model computation and defeats the original purpose of MAE efficient computation. 
 Second, class token plays a crucial role in the fine-tuning process of MAE. However, the class token of MAE and its improvement works are not involved in the task during the pre-training phase. 
 Therefore, it is necessary to design a contrastive learning task that keeps the MAE computation efficient while involving the class token.

 Given an input image ${x} \in {\mathbb{R}^{H \times W \times 3}}$, we apply different intensities of data-augment to it to obtain $x_s$ and $x_w$. 
 We encode only the set of visible tokens for these two images to obtain ${z}_{vis}^{s}$, ${z}_{vis}^{w}$. This greatly improves the computational efficiency of SDMAE. 
 Then our decoder recovers the full set of tokens for both images. This process can be depicted as:
 \begin{equation}
   \begin{split}
         &{{z}}_{{{all}}}^{{w}} = \left[{z}_{vis}^{{w}},{{z}}_{{m}}^{{w}}\right],\\
         &{{z}}_{{{all}}}^{{s}} = \left[{{z}}_{vis}^{{s}},{{z}}_{{m}}^{{s}}\right],\\
         &{{z}_{all}^w}' = Dec({{z}}_{{{all}}}^{{w}} + {E_{{{pos}}}}),\\
         &{{{z}_{all}^s}'} = Dec({{z}}_{{{all}}}^{{s}} + {E_{{{pos}}}}),
 \end{split}
   \label{eq3}
 \end{equation}
 where ${{z}}_{{m}}^{{s}}$ and ${{z}}_{{m}}^{{w}}$ are the learnable mask tokens for the strong and weak data augmented images, respectively. 
 $\left[.\right]$ represents the merge operation, and $Dec\left(.\right)$ is the decoder of SDMAE.

 Next ${{z}_{{{all}}}^{{w}}}'$ and ${{z}_{{{all}}}^{{s}}}'$ are encoded again by two small encoders with two transformer blocks ${{{e}}^{{q}}}$ and ${{{e}}^{{k}}}$, 
 and ${{{e}}^{{k}}}$ is the momentum encoder for ${{{e}}^{{q}}}$. This can be specified as:
 \begin{equation}
      {\theta _k} = a{\theta _k} + (1 - a){\theta _q}
   \label{eq4}
 \end{equation}
 where ${\theta _k}$ is the parameter of ${{{e}}^{{k}}}$ and ${\theta _q}$ is the parameter of ${{{e}}^{{q}}}$. $a$ is the momentum update weight. 
 To implement the asymmetric structure design for contrastive learning, we only project one of the above codes. 
 Our projection layer consists of two expansive fully connected layers. 
 Finally, we extract class token from the output of the encoder and projection layer of the two images respectively for contrastive training. This process is calculated as:
 \begin{equation}
   \begin{split}
         &{q_w} = proj({{{e}}^{{q}}}({{{z}}_{{{all}}}^{{w}}}')[:1]),\\
         &{k_w} = {{{e}}^{{k}}}({{{z}}_{{{all}}}^{{w}}}')[:1],\\
         &{q_s} = proj({{{e}}^{{k}}}({{{z}}_{{{all}}}^{{s}}}')[:1]),\\
         &{k_s} = {{{e}}^{{k}}}({{{z}}_{{{all}}}^{{s}}}')[:1],
 \end{split}
   \label{eq5}
 \end{equation}
 where $proj(.)$ denotes projector and $[:1]$ represents the extract class token operation. Our contrastive losses can be written as: 
 \begin{equation}
   \begin{split}
         &{L_{{{ctr}}}} =  -(\log \frac{{\exp ({q_s}.{k_w}/\tau )}}{{\exp ({q_s}.{k_w}/\tau ) + \sum\limits_{{z_k^ -}} {\exp ({q_s}.k_w^ - /\tau )} }}+\\
         &\log \frac{{\exp ({q_w}{k_s}/\tau )}}{{\exp ({q_w}.{k_s}/\tau ) + \sum\limits_{{z_k^ -}} {\exp ({q_w}{k_s^ -}/\tau )} }})
 \end{split}
   \label{eq6}
 \end{equation}
 Where $\tau $ is a temperature hyper-parameter.

 \subsection{Loss Function}

 We select mask area tokens ${{y' }}_{{m}}^{{s}}$ and ${{y' }}_{{m}}^{{w}}$ from ${{{z}}_{{{all}}}^{{s}}}'$ and ${{{z}}_{{{all}}}^{{w}}}'$ 
 respectively to predict the original set of masked tokens ${{x}}_{{m}}^{{s}}$ and ${{x}}_{{m}}^{{w}}$. 
 The whole reconstruction loss can be expressed as:
 \begin{equation}
   \begin{split}
         &Y_m^s = Norm({{x}}_{{m}}^{{s}}),\\
         &Y_m^{{w}} = Norm({{x}}_{{m}}^{{w}}),\\
         &{L_{{{con}}}} = MSE({{y' }}_{{m}}^{{s}},Y_m^s) + MSE({{y' }}_{{m}}^{{w}},Y_m^{{w}}),
 \end{split}
   \label{eq7}
 \end{equation}
 The total loss of SDMAE is recorded as:
 \begin{equation}
      {L_{total}} = {L_{{{con}}}} + {\lambda _l}{L_{loc}} + {\lambda _c}{L_{ctr}},
   \label{eq8}
 \end{equation}
 Where ${\lambda _l}$ and ${\lambda _c}$ are hyperparameters for location prediction loss and contrastive loss respectively.  Unless otherwise specified, in the following experiments, We default ${\lambda _c}=0.1 , {\lambda _l}=1$.

 \section{Experiments}
 
 We conduct image classification experiments on several standard small datasets. 
 In addition, we also perform diagnostic work on image-sparse medical datasets to validate the performance of SDMAE in practical applications with small datasets. 
 We next compare SDMAE with current state-of-the-art algorithms in these tasks. We then conduct an ablation study on the core designs of SDMAE.

 \subsection{Image Classification on small datasets}

 \paragraph{Settings.} For small dataset image classification experiments, we evaluate SDMAE on four datasets, Tiny-ImageNet (T-IN)\cite{tavanaei2020embedded}, CIFAR-100\cite{krizhevsky2009learning}, 
 CIFAR-10\cite{krizhevsky2009learning} and SVHN\cite{liao2015competitive}.
 Please see Appendix \textcolor{red}{A} for a detailed description of the dataset. 
 The top-1 accuracy on four datasets is reported. We follow with the detailed settings for the pre-training and fine-tuning phases:
 \begin{itemize}
   \item pre-training. We use AdamW\cite{loshchilov2017decoupled} with a base learning rate of 1e-3 to pre-train SDMAE for 300 epochs. The learning rate scheduler adopts cosine decay with 40 epochs of linear warm-up. SDMAE employs 64 batches of 224 resolution images as input. The mask ratio and weight decay we set to 0.75 and 0.05, respectively. For the model architecture we choose the standard configuration ViT-Base as the encoder. For the data augmentation scheme, we follow MoCo v3\cite{chen2021empirical}.
   \item fine-tuning. We use 100 epochs and 20 epochs for linear warm-up fine-tuning scheme. Furthermore, we have added a drop path rate of 0.1. We followed the data augmentation scheme of the MAE\cite{he2022masked} fine-tuning phase, and the rest of the settings were kept consistent with the pre-training phase.
\end{itemize}

\begin{table}[ht]
   \centering
   \footnotesize
   \setlength{\tabcolsep}{1.0mm}
   \begin{tabular}{l|c|c|c|c|c}   
      Method            & \#FLOPs(M)        & CIFAR-10           &CIFAR-100           & SVHN              & T-IN \\
      \hline
      ResNet56\cite{he2016deep}          & 506.2             & 95.7              & 76.36             & 97.73             & 58.77 \\
      ResNet110\cite{he2016deep}        & 1020              & 96.37             & 79.86             & 97.85             & 62.96 \\
      EffecientNet B0\cite{tan2019efficientnet}   & 123.9             & 94.66             & 76.04             & 97.22             & 66.79 \\
      MobileNet V3\cite{howard2019searching}      & 233.2             & 81.52             & 51.74             & 91.48             & 36.56 \\
      ShuffleNet V2\cite{ma2018shufflenet}     & 591               & 82.21             & 51.44             & 91.88             & 35.46 \\
      \hline
      ViT-Tiny\cite{dosovitskiy2020image}          & 189.8             & 93.58             & 73.81             & 97.82             & 57.07 \\
      ViT-Base\cite{dosovitskiy2020image}          & 16863.6           & 91.91             & 67.52             & 97.8              & 56.52 \\
      SL-ViT\cite{lee2021vision}            & 199.2             & 94.53             & 76.92             & 97.79             & 61.07 \\
      SL-Cait\cite{lee2021vision}           & 623.3             & 95.8              & 80.3              & 98.2              & 67.1 \\
      SL-PiT w/o ${\rm{S}}_{pool}$\cite{lee2021vision}& 280.4             & 94.96             & 77.08             & 97.94             & 60.31 \\
      SL-PiT w/ ${\rm{S}}_{pool}$\cite{lee2021vision}& 322.9             & 95.88             & 79                & 97.93             & 62.91 \\
      SL-Swin w/o ${\rm{S}}_{pool}$\cite{lee2021vision}& 247               & 95.3              & 78.13             & 97.88             & 62.7 \\
      SL-Swin w/ ${\rm{S}}_{pool}$\cite{lee2021vision}& 284.9             & 95.93             & 79.99             & 97.92             & 64.95 \\
      CCT-7/3×1\cite{hassani2021escaping}         & 1190              & 96.53             & 80.92             & -                 & -     \\
      Drloc-ViT\cite{liu2021efficient}         & 189.8             & 81                & 58.29             & 94.02             & 42.33 \\
      Drloc-Swin\cite{liu2021efficient}        & 239.6             & 83.89             & 66.23             & 94.23             & 48.66 \\
      Drloc-CaiT\cite{liu2021efficient}        & 472               & 82.2              & 56.32             & 19.59             & 45.95 \\
      SSL-ViT\cite{gani2022train}          & 353.1             & 96.41             & 79.15             & 98.03             & 63.36 \\
      SSL-Swin\cite{gani2022train}          & 479.2             & 96.18             & 80.95             & 98.01             & 65.13 \\
      SSL-CaiT\cite{gani2022train}          & 944.1             & 96.42             & 80.79             & 98.18             & 67.46 \\
      \rowcolor{mygray}
      Ours              & 20164.7           & 96.57             & 82                & 98.3              & 72.24 \\
     \end{tabular}
   \caption{Top-1 accuracy comparison of CNNs and ViTs on small-size datasets. We refer to \cite{gani2022train} and its variants as SSL-ViT, SSL-Swin and SSL-CaiT.}
\label{tab1}
\end{table}

\paragraph{Comparison with CNNs and ViTs for small datasets.} 

Table~\ref{tab1} shows the results of SDMAE compared with ViTs models for small datasets and advanced CNNs. 
SDMAE successfully helps the standard configuration of ViT models to significantly outperform CNNs on four small datasets. 
SDMAE also achieves state-of-the-art performance compared to current works on ViTs trained on small datasets. +1.08\% for SDMAE(82\%) over CCT\cite{hassani2021escaping} on CIFAR-100, 
+5.1\% for SDMAE(72.2\%) over SL-CaiT\cite{touvron2021going} on Tiny-ImageNet, +0.1\% for SDMAE(98.3\%) over SL-CaiT on SVNH and +0.8\% for SDMAE(96.6\%) over SL-CaiT on CIFAR-10.

\begin{table}[ht]
   \centering
   \footnotesize
   \setlength{\tabcolsep}{1.0mm}
   \begin{tabular}{l|c|c|c|c|c}   

      Method            & pre-train epochs & CIFAR-10            & CIFAR-100          & SVHN              & T-IN \\
      \hline
      MAE\cite{he2022masked}               & 300               & 93.41             & 75.15             & 97.66             & 62.95 \\
      i-MAE\cite{zhang2022mae}             & 1000              & 92.34             & 69.5              & -                 & 61.13 \\
      ConvMAE\cite{gao2022convmae}           & 300               & 90.9              & 73.19             & 96.94             & 62.26 \\
      SimMIM\cite{xie2022simmim}            & 300               & 68.17             & 66.4              & 96.02             & 49.86 \\
      BEiT\cite{bao2021beit}              & 300               & 82.82             & 70.29             & 81.23             & 48.44 \\
      \rowcolor{mygray}
      Ours              & 300               & 96.57             & 82                & 98.3              & 72.24 \\

     \end{tabular}
   \caption{Top-1 accuracy comparison with different MIMs on small-size datasets.}
\label{tab2}
\end{table}

\begin{figure}[ht]
   \centering
   \includegraphics[width=1\linewidth]{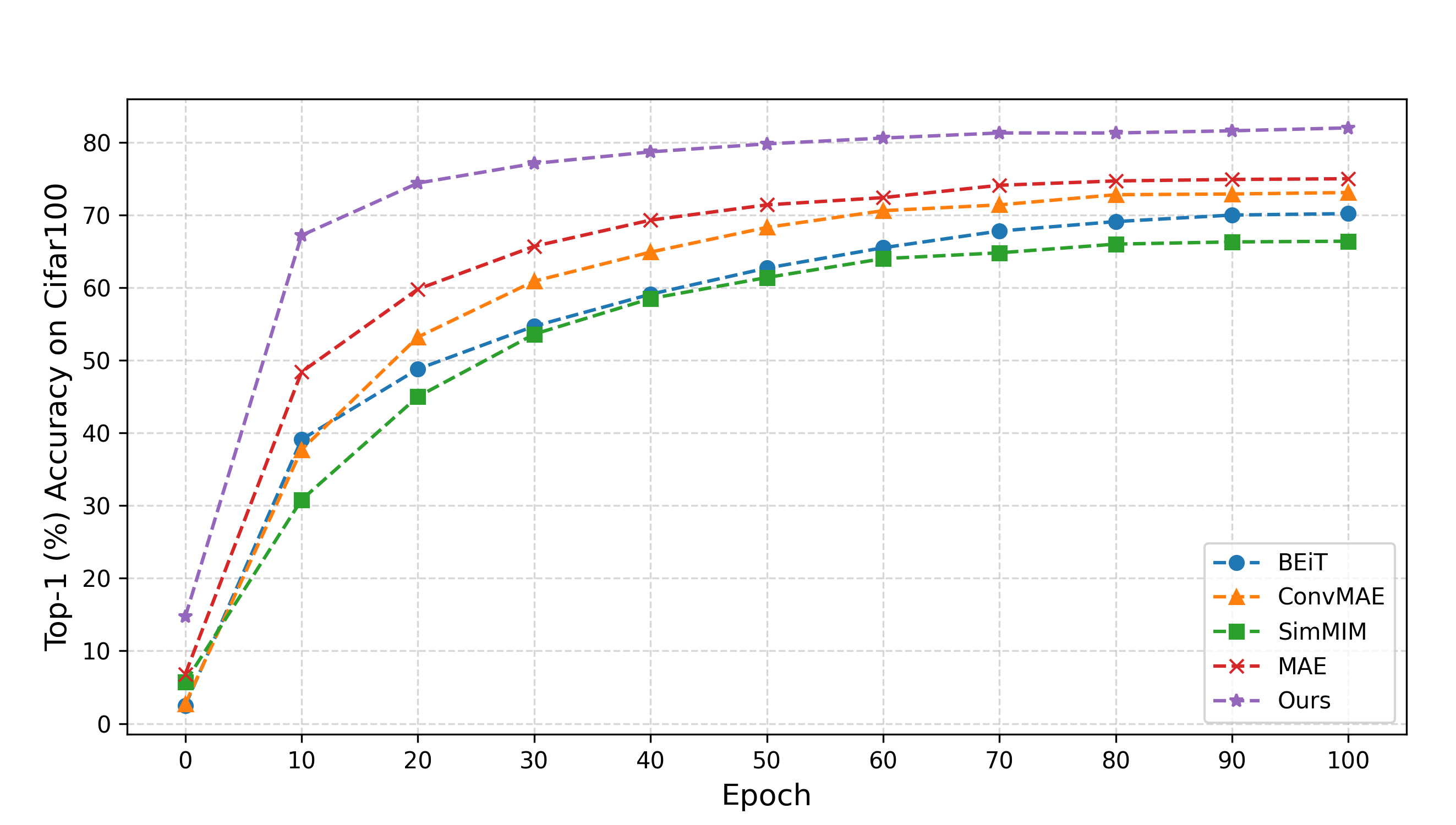}
    \caption{Comparison of our method with the fine-tuning results of several MIMs on CIFAR-100.}
    \label{fig7}
 \end{figure}

\paragraph{Comparison with MIM for small datasets.} We show the results of comparing SDMAE with the currently popular MIM model on four small datasets in Table~\ref{tab2}. 
Compared with several MIMs, our SDMAE shows superior performance on small datasets. +9.29\% for SDMAE(72.2\%) over MAE\cite{he2022masked} and +9.98\% for SDMAE over ConvMAE\cite{gao2022convmae} on Tiny-ImageNet. 
This suggests that our SDMAE is more suitable for training standard configuration ViT models on small datasets than other MIMs. 
Figure~\ref{fig7} shows training efficiency of SDMAE in comparison to other MIMs during the fine-tuning stage.

 \subsection{Medical Image Diagnosis with small datasets}

\begin{table}[ht]
   \centering
   \footnotesize
   \setlength{\tabcolsep}{1.0mm}
   \begin{tabular}{l|c|c|c}

      Method            & Resolution & APTOS 2019 & COVID-19 \\
      \hline
      ResNet110\cite{he2016deep}         & ${64}^2$                & 71.85             & 56.5 \\
      EffecientNet B0\cite{tan2019efficientnet}   & ${64}^2$          & 73.49             & 53 \\
      MAE\cite{he2022masked}               & ${224}^2$               & 82.79             & 60.5 \\
      ConvMAE\cite{gao2022convmae}           & ${224}^2$               & 80.6              & 58 \\
      SSL-ViT\cite{gani2022train}               & ${64}^2$                & 56.28             & 60.5 \\
      Drloc-ViT\cite{liu2021efficient}          & ${224}^2$               & 74.59             & 59.5 \\
      DIRA\cite{haghighi2022dira}              & ${224}^2$               & 58.19             & 58.5 \\
      Medical MAE\cite{xiao2022delving}       & ${224}^2$               & 76.77             & 60 \\
      \rowcolor{mygray}
      Ours              & ${224}^2$                 & 83.06             & 61 \\
  
     \end{tabular}
   \caption{Effectiveness of SDMAE on medical datasets including only few images. We refer to \cite{gani2022train} as SSL-ViT.}
\label{tab3}
\end{table}

\paragraph{Settings.}To validate the performance of SDMAE in real-world applications with small datasets, we apply it to medical diagnostic work. 
We selected APTOS 2019\cite{kaggle} and COVID-19\cite{zhao2020covid}, which contain fewer images, for evaluation. APTOS 2019 contains only 3662 images. 
For COVID-19, we use only 546 images for training. For more details on the use of the medical datasets, please see Appendix \textcolor{red}{A}. 
The pre-training and fine-tuning phase settings are kept consistent with the image classification experiments.

\begin{figure}[ht]
   \centering
   \includegraphics[width=1\linewidth]{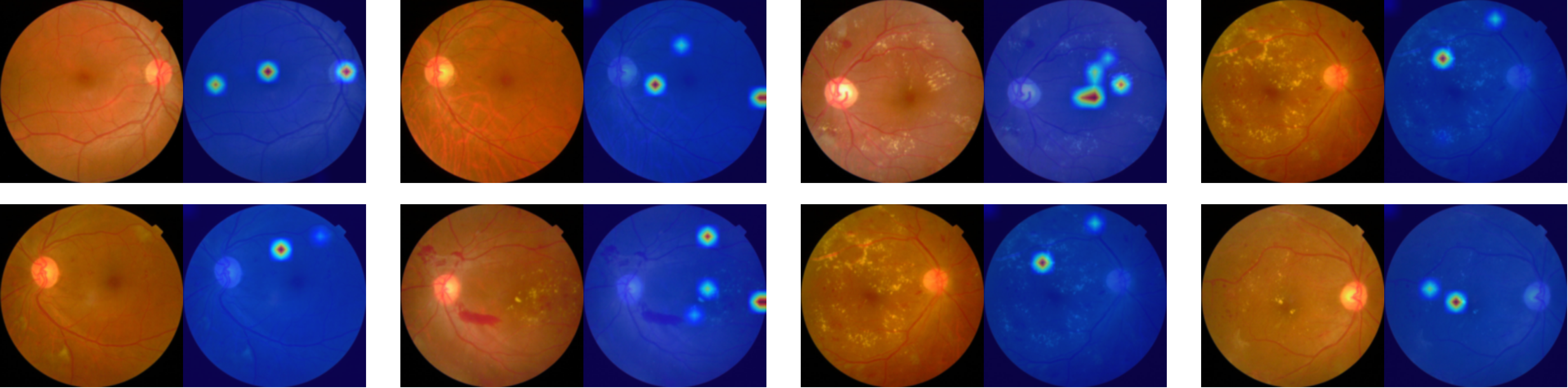}
    \caption{The last layer attention visualization results of our method on the APTOS 2019 dataset. 
    The left side is the input image, and the right side is the corresponding attention visualization result.}
    \label{fig8}
 \end{figure}

 \paragraph{Results.}We quantitatively and qualitatively demonstrate the results of applying SDMAE to a small dataset of medical diagnoses. 
Table~\ref{tab3} shows that SDMAE shows state-of-the-art results compared to advanced medical diagnostic algorithms, MIMs and CNNs, and small datasets ViTs. 
 This shows that our SDMAE can indeed be effective in real-world applications with small datasets. 
 We use the attention visualization in Figure~\ref{fig8} to show the regions that SDMAE focuses on in the medical datasets. 

 \subsection{Ablation Study}
 In this section, we investigate the optimal decoder configuration for SDMAE and focus on verifying the effectiveness of the location prediction task as well as the comparative task in SDMAE. 
 We mainly used the image classification task on Tiny-ImageNet for their ablation study. 
 
 \begin{figure}[ht]
   \centering
   \includegraphics[width=1\linewidth]{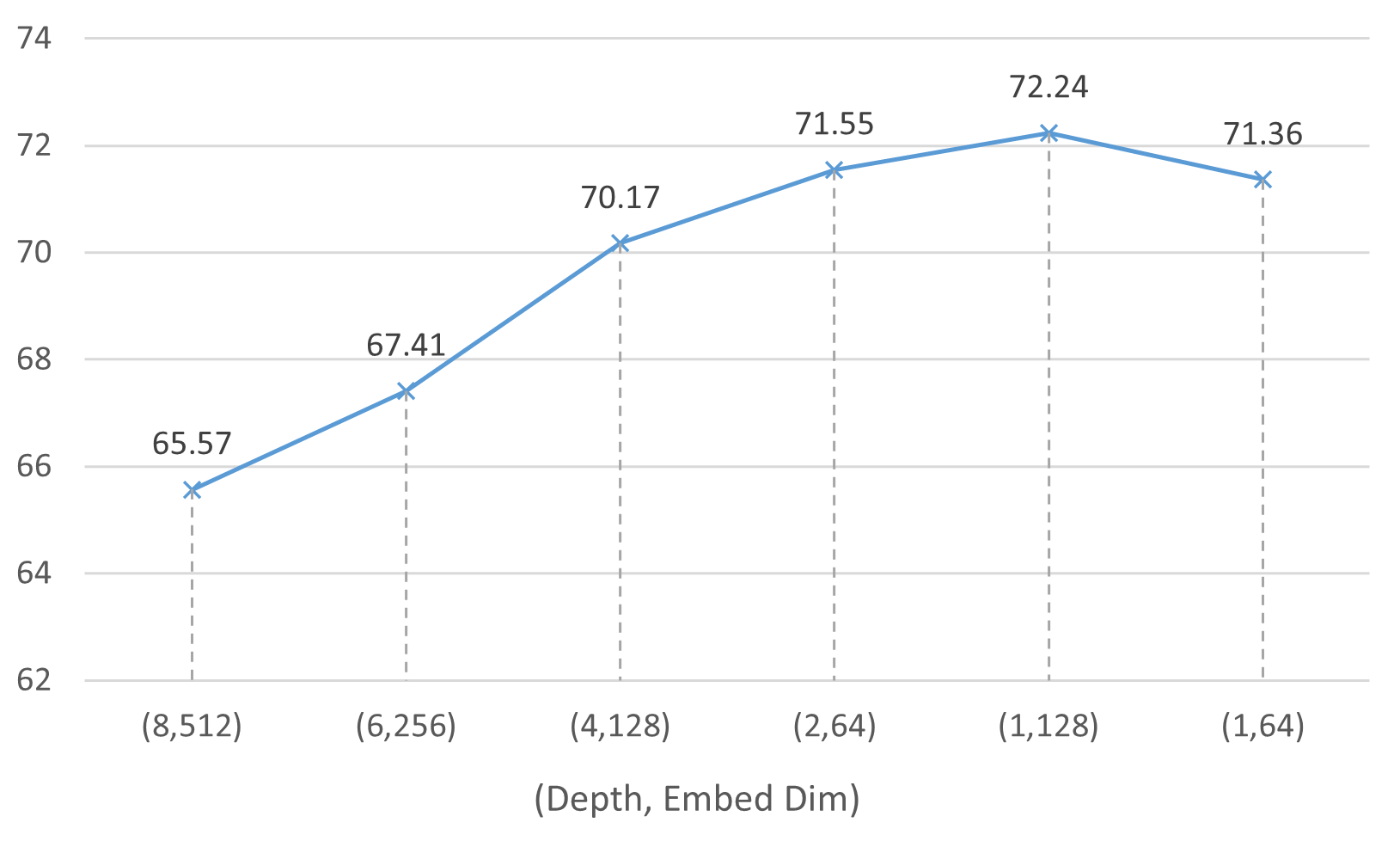}
    \caption{Influence of a series of decoder configurations on SDMAE fine-tuning results on Tiny-ImageNet.}
    \label{fig9}
 \end{figure}
\begin{figure*}[t]
   \centering
   \includegraphics[width=0.9\linewidth]{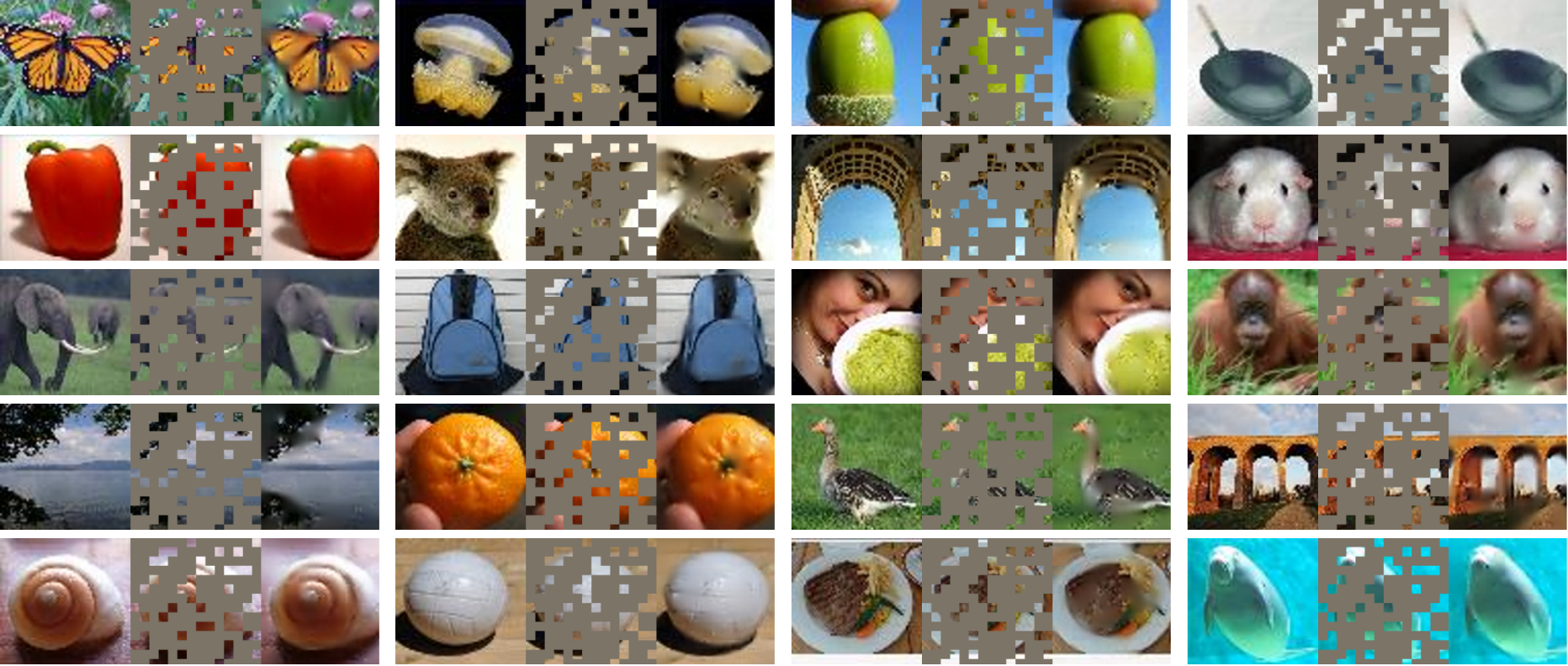}
    \caption{Image prediction results of SDMAE. For each triplet, we show the input image (left), the masked image (middle), and  our SDMAE prediction result (right). We adopts 75\% masking ratio.}
    \label{fig10}
 \end{figure*}
 \paragraph{Decoder configuration.}Based on the phenomena we observe in Section \textcolor{red}{3.2}, we test the classification results on Tiny-ImageNet for a range of combinations of depth and embedding dimensions. 
 Since a decoder with embedding dimension 32 can severely damage the predicted image, we do not investigate the decoder configuration with embedding dimension 32 anymore. 
 Figure~\ref{fig9} shows the experimental results. 
 We can observe that SDMAE presents the best results when using a decoder configuration with a depth of 1 embedding dimension of 128.

   \begin{table}[ht]
      \footnotesize
      \setlength{\tabcolsep}{8pt}
      \begin{tabular}{cc|c}
   
         Location Prediction Task &  Contrastive Task & Top-1 Acc\\
   \toprule
    \xmark & \xmark & 70.59 \\
    \xmark & \cmark & 70.94 \\
    \cmark & \xmark & 71.22 \\
    \cmark & \cmark & 72.24 \\ 
   \end{tabular}
      \caption{ Ablation results for the location prediction task and the contrastive task in SDMAE.}
   \label{tab4}
   \end{table}

\paragraph{Location Prediction Task.}We report the ablation results for the location prediction task in Table~\ref{tab4}. 
It can be found that when SDMAE does not perform the location prediction task, the performance of the standard configuration ViT on Tiny-ImageNet decreases by 1.3\%. 
This shows that the location prediction task is essential to help the ViT in SDMAE capture local spatial information. 
Therefore, we conclude that it is necessary for ViT to design tasks related to local information in small datasets training.

\paragraph{Contrastive Task.}Table~\ref{tab4} shows the ablation results of the contrastive task in SDMAE. 
We see that when we cancel the SDMAE comparison task, its top-1 accuracy on Tiny-ImageNet decreased significantly. 
This is due to the cancellation of the contrastive task, which makes SDMAE unable to learn the invariance of CNNs, 
and the lack of specific task constraints on class token in the pre-training stage. 

\subsection{Visualization}

To demonstrate the image prediction performance of SDMAE, we show the prediction results on Tiny-ImageNet in Figure~\ref{fig10}. We can see that even though SDMAE weakens the decoder, it still successfully predicts the image. See Appendix \textcolor{red}{B} for more forecast results.

\section{Conclusion}

In this paper, we propose a method to efficiently train ViT on small datasets by MAE, which is called small dataset masked autoencoders (SDMAE). 
We first solve the over-fitting issue of MAE decoder on small data sets experimentally and present a decoder configuration for SDMAE that is suitable for training on small datasets.
Since SDMAE does not require changing the configuration of the ViT model, it achieves unification of ViT training on both large and small datasets.  
Secondly, we also propose a location prediction task as well as a contrastive task to introduce the properties of localization and invariance of CNNs. 
Our contrastive task design successfully reduces the computational cost of MAE improvement work combined with contrastive learning and considers the training of class token in the pre-training phase. 
Numerous experiments have shown that SDMAE significantly improves the quality of learning representations on small datasets with standard configuration of ViT. 
Our SDMAE achieves state-of-the-art performance on both several standard small datasets and image-sparse medical datasets. 
This indicates that SDMAE is an effective means to solve transformer data-hungry.

\section*{Acknowledge} This work was supported by Public-welfare Technology Application Research of Zhejiang Province in China under Grant LGG22F020032, and Key Research and Development Project of Zhejiang Province in China under Grant 2021C03137.

{\small
\bibliographystyle{ieee_fullname}
\bibliography{egbib}

\begin{thebibliography}{10}\itemsep=-1pt

\bibitem{bao2021beit}
Hangbo Bao, Li Dong, and Furu Wei.
\newblock Beit: Bert pre-training of image transformers.
\newblock {\em arXiv preprint arXiv:2106.08254}, 2021.

\bibitem{bauer1999empirical}
Eric Bauer and Ron Kohavi.
\newblock An empirical comparison of voting classification algorithms: Bagging,
  boosting, and variants.
\newblock {\em Machine learning}, 36(1):105--139, 1999.

\bibitem{chen2021regionvit}
Chun-Fu Chen, Rameswar Panda, and Quanfu Fan.
\newblock Regionvit: Regional-to-local attention for vision transformers.
\newblock {\em arXiv preprint arXiv:2106.02689}, 2021.

\bibitem{chen2021crossvit}
Chun-Fu~Richard Chen, Quanfu Fan, and Rameswar Panda.
\newblock Crossvit: Cross-attention multi-scale vision transformer for image
  classification.
\newblock In {\em Proceedings of the IEEE/CVF international conference on
  computer vision}, pages 357--366, 2021.

\bibitem{chen2022context}
Xiaokang Chen, Mingyu Ding, Xiaodi Wang, Ying Xin, Shentong Mo, Yunhao Wang,
  Shumin Han, Ping Luo, Gang Zeng, and Jingdong Wang.
\newblock Context autoencoder for self-supervised representation learning.
\newblock {\em arXiv preprint arXiv:2202.03026}, 2022.

\bibitem{chen2021empirical}
Xinlei Chen, Saining Xie, and Kaiming He.
\newblock An empirical study of training self-supervised vision transformers.
\newblock In {\em Proceedings of the IEEE/CVF International Conference on
  Computer Vision}, pages 9640--9649, 2021.

\bibitem{chen2022sdae}
Yabo Chen, Yuchen Liu, Dongsheng Jiang, Xiaopeng Zhang, Wenrui Dai, Hongkai
  Xiong, and Qi Tian.
\newblock Sdae: Self-distillated masked autoencoder.
\newblock In {\em European Conference on Computer Vision}, pages 108--124.
  Springer, 2022.

\bibitem{chu2021twins}
Xiangxiang Chu, Zhi Tian, Yuqing Wang, Bo Zhang, Haibing Ren, Xiaolin Wei,
  Huaxia Xia, and Chunhua Shen.
\newblock Twins: Revisiting the design of spatial attention in vision
  transformers.
\newblock {\em Advances in Neural Information Processing Systems},
  34:9355--9366, 2021.

\bibitem{deng2009imagenet}
Jia Deng, Wei Dong, Richard Socher, Li-Jia Li, Kai Li, and Li Fei-Fei.
\newblock Imagenet: A large-scale hierarchical image database.
\newblock In {\em 2009 IEEE conference on computer vision and pattern
  recognition}, pages 248--255. Ieee, 2009.

\bibitem{devlin2018bert}
Jacob Devlin, Ming-Wei Chang, Kenton Lee, and Kristina Toutanova.
\newblock Bert: Pre-training of deep bidirectional transformers for language
  understanding.
\newblock {\em arXiv preprint arXiv:1810.04805}, 2018.

\bibitem{ding2022decoupling}
Jian Ding, Nan Xue, Gui-Song Xia, and Dengxin Dai.
\newblock Decoupling zero-shot semantic segmentation.
\newblock In {\em Proceedings of the IEEE/CVF Conference on Computer Vision and
  Pattern Recognition}, pages 11583--11592, 2022.

\bibitem{dong2022bootstrapped}
Xiaoyi Dong, Jianmin Bao, Ting Zhang, Dongdong Chen, Weiming Zhang, Lu Yuan,
  Dong Chen, Fang Wen, and Nenghai Yu.
\newblock Bootstrapped masked autoencoders for vision bert pretraining.
\newblock In {\em European Conference on Computer Vision}, pages 247--264.
  Springer, 2022.

\bibitem{dosovitskiy2020image}
Alexey Dosovitskiy, Lucas Beyer, Alexander Kolesnikov, Dirk Weissenborn,
  Xiaohua Zhai, Thomas Unterthiner, Mostafa Dehghani, Matthias Minderer, Georg
  Heigold, Sylvain Gelly, et~al.
\newblock An image is worth 16x16 words: Transformers for image recognition at
  scale.
\newblock {\em arXiv preprint arXiv:2010.11929}, 2020.

\bibitem{el2021large}
Alaaeldin El-Nouby, Gautier Izacard, Hugo Touvron, Ivan Laptev, Herv{\'e}
  Jegou, and Edouard Grave.
\newblock Are large-scale datasets necessary for self-supervised pre-training?
\newblock {\em arXiv preprint arXiv:2112.10740}, 2021.

\bibitem{gani2022train}
Hanan Gani, Muzammal Naseer, and Mohammad Yaqub.
\newblock How to train vision transformer on small-scale datasets?
\newblock {\em arXiv preprint arXiv:2210.07240}, 2022.

\bibitem{gao2022convmae}
Peng Gao, Teli Ma, Hongsheng Li, Jifeng Dai, and Yu Qiao.
\newblock Convmae: Masked convolution meets masked autoencoders.
\newblock {\em arXiv preprint arXiv:2205.03892}, 2022.

\bibitem{graham2021levit}
Benjamin Graham, Alaaeldin El-Nouby, Hugo Touvron, Pierre Stock, Armand Joulin,
  Herv{\'e} J{\'e}gou, and Matthijs Douze.
\newblock Levit: a vision transformer in convnet's clothing for faster
  inference.
\newblock In {\em Proceedings of the IEEE/CVF international conference on
  computer vision}, pages 12259--12269, 2021.

\bibitem{guo2022simt}
Xiaoqing Guo, Jie Liu, Tongliang Liu, and Yixuan Yuan.
\newblock Simt: Handling open-set noise for domain adaptive semantic
  segmentation.
\newblock In {\em Proceedings of the IEEE/CVF Conference on Computer Vision and
  Pattern Recognition}, pages 7032--7041, 2022.

\bibitem{haghighi2022dira}
Fatemeh Haghighi, Mohammad Reza~Hosseinzadeh Taher, Michael~B Gotway, and
  Jianming Liang.
\newblock Dira: Discriminative, restorative, and adversarial learning for
  self-supervised medical image analysis.
\newblock In {\em Proceedings of the IEEE/CVF Conference on Computer Vision and
  Pattern Recognition}, pages 20824--20834, 2022.

\bibitem{hassani2021escaping}
Ali Hassani, Steven Walton, Nikhil Shah, Abulikemu Abuduweili, Jiachen Li, and
  Humphrey Shi.
\newblock Escaping the big data paradigm with compact transformers.
\newblock {\em arXiv preprint arXiv:2104.05704}, 2021.

\bibitem{he2022masked}
Kaiming He, Xinlei Chen, Saining Xie, Yanghao Li, Piotr Doll{\'a}r, and Ross
  Girshick.
\newblock Masked autoencoders are scalable vision learners.
\newblock In {\em Proceedings of the IEEE/CVF Conference on Computer Vision and
  Pattern Recognition}, pages 16000--16009, 2022.

\bibitem{he2016deep}
Kaiming He, Xiangyu Zhang, Shaoqing Ren, and Jian Sun.
\newblock Deep residual learning for image recognition.
\newblock In {\em Proceedings of the IEEE conference on computer vision and
  pattern recognition}, pages 770--778, 2016.

\bibitem{heo2021rethinking}
Byeongho Heo, Sangdoo Yun, Dongyoon Han, Sanghyuk Chun, Junsuk Choe, and
  Seong~Joon Oh.
\newblock Rethinking spatial dimensions of vision transformers.
\newblock In {\em Proceedings of the IEEE/CVF International Conference on
  Computer Vision}, pages 11936--11945, 2021.

\bibitem{howard2019searching}
Andrew Howard, Mark Sandler, Grace Chu, Liang-Chieh Chen, Bo Chen, Mingxing
  Tan, Weijun Wang, Yukun Zhu, Ruoming Pang, Vijay Vasudevan, et~al.
\newblock Searching for mobilenetv3.
\newblock In {\em Proceedings of the IEEE/CVF international conference on
  computer vision}, pages 1314--1324, 2019.

\bibitem{huang2017densely}
Gao Huang, Zhuang Liu, Laurens Van Der~Maaten, and Kilian~Q Weinberger.
\newblock Densely connected convolutional networks.
\newblock In {\em Proceedings of the IEEE conference on computer vision and
  pattern recognition}, pages 4700--4708, 2017.

\bibitem{huang2022contrastive}
Zhicheng Huang, Xiaojie Jin, Chengze Lu, Qibin Hou, Ming-Ming Cheng, Dongmei
  Fu, Xiaohui Shen, and Jiashi Feng.
\newblock Contrastive masked autoencoders are stronger vision learners.
\newblock {\em arXiv preprint arXiv:2207.13532}, 2022.

\bibitem{jin2022you}
Zhenchao Jin, Dongdong Yu, Luchuan Song, Zehuan Yuan, and Lequan Yu.
\newblock You should look at all objects.
\newblock In {\em European Conference on Computer Vision}, pages 332--349.
  Springer, 2022.

\bibitem{kaggle}
Kaggle.
\newblock Aptos 2019 blindness detection, 2019.

\bibitem{kim2022restr}
Namyup Kim, Dongwon Kim, Cuiling Lan, Wenjun Zeng, and Suha Kwak.
\newblock Restr: Convolution-free referring image segmentation using
  transformers.
\newblock In {\em Proceedings of the IEEE/CVF Conference on Computer Vision and
  Pattern Recognition}, pages 18145--18154, 2022.

\bibitem{krizhevsky2009learning}
Alex Krizhevsky, Geoffrey Hinton, et~al.
\newblock Learning multiple layers of features from tiny images.
\newblock 2009.

\bibitem{krizhevsky2017imagenet}
Alex Krizhevsky, Ilya Sutskever, and Geoffrey~E Hinton.
\newblock Imagenet classification with deep convolutional neural networks.
\newblock {\em Communications of the ACM}, 60(6):84--90, 2017.

\bibitem{lecun1998gradient}
Yann LeCun, L{\'e}on Bottou, Yoshua Bengio, and Patrick Haffner.
\newblock Gradient-based learning applied to document recognition.
\newblock {\em Proceedings of the IEEE}, 86(11):2278--2324, 1998.

\bibitem{lee2021vision}
Seung~Hoon Lee, Seunghyun Lee, and Byung~Cheol Song.
\newblock Vision transformer for small-size datasets.
\newblock {\em arXiv preprint arXiv:2112.13492}, 2021.

\bibitem{li2022deep}
Liulei Li, Tianfei Zhou, Wenguan Wang, Jianwu Li, and Yi Yang.
\newblock Deep hierarchical semantic segmentation.
\newblock In {\em Proceedings of the IEEE/CVF Conference on Computer Vision and
  Pattern Recognition}, pages 1246--1257, 2022.

\bibitem{li2022oriented}
Wentong Li, Yijie Chen, Kaixuan Hu, and Jianke Zhu.
\newblock Oriented reppoints for aerial object detection.
\newblock In {\em Proceedings of the IEEE/CVF Conference on Computer Vision and
  Pattern Recognition}, pages 1829--1838, 2022.

\bibitem{li2022sepvit}
Wei Li, Xing Wang, Xin Xia, Jie Wu, Xuefeng Xiao, Min Zheng, and Shiping Wen.
\newblock Sepvit: Separable vision transformer.
\newblock {\em arXiv preprint arXiv:2203.15380}, 2022.

\bibitem{liao2015competitive}
Zhibin Liao and Gustavo Carneiro.
\newblock Competitive multi-scale convolution.
\newblock {\em arXiv preprint arXiv:1511.05635}, 2015.

\bibitem{liu2021efficient}
Yahui Liu, Enver Sangineto, Wei Bi, Nicu Sebe, Bruno Lepri, and Marco Nadai.
\newblock Efficient training of visual transformers with small datasets.
\newblock {\em Advances in Neural Information Processing Systems},
  34:23818--23830, 2021.

\bibitem{liu2021swin}
Ze Liu, Yutong Lin, Yue Cao, Han Hu, Yixuan Wei, Zheng Zhang, Stephen Lin, and
  Baining Guo.
\newblock Swin transformer: Hierarchical vision transformer using shifted
  windows.
\newblock In {\em Proceedings of the IEEE/CVF International Conference on
  Computer Vision}, pages 10012--10022, 2021.

\bibitem{liu2022convnet}
Zhuang Liu, Hanzi Mao, Chao-Yuan Wu, Christoph Feichtenhofer, Trevor Darrell,
  and Saining Xie.
\newblock A convnet for the 2020s.
\newblock In {\em Proceedings of the IEEE/CVF Conference on Computer Vision and
  Pattern Recognition}, pages 11976--11986, 2022.

\bibitem{loshchilov2017decoupled}
Ilya Loshchilov and Frank Hutter.
\newblock Decoupled weight decay regularization.
\newblock {\em arXiv preprint arXiv:1711.05101}, 2017.

\bibitem{ma2018shufflenet}
Ningning Ma, Xiangyu Zhang, Hai-Tao Zheng, and Jian Sun.
\newblock Shufflenet v2: Practical guidelines for efficient cnn architecture
  design.
\newblock In {\em Proceedings of the European conference on computer vision
  (ECCV)}, pages 116--131, 2018.

\bibitem{matsoukas2021time}
Christos Matsoukas, Johan~Fredin Haslum, Magnus S{\"o}derberg, and Kevin Smith.
\newblock Is it time to replace cnns with transformers for medical images?
\newblock {\em arXiv preprint arXiv:2108.09038}, 2021.

\bibitem{simonyan2014very}
Karen Simonyan and Andrew Zisserman.
\newblock Very deep convolutional networks for large-scale image recognition.
\newblock {\em arXiv preprint arXiv:1409.1556}, 2014.

\bibitem{sun2017revisiting}
Chen Sun, Abhinav Shrivastava, Saurabh Singh, and Abhinav Gupta.
\newblock Revisiting unreasonable effectiveness of data in deep learning era.
\newblock In {\em Proceedings of the IEEE international conference on computer
  vision}, pages 843--852, 2017.

\bibitem{szegedy2015going}
Christian Szegedy, Wei Liu, Yangqing Jia, Pierre Sermanet, Scott Reed, Dragomir
  Anguelov, Dumitru Erhan, Vincent Vanhoucke, and Andrew Rabinovich.
\newblock Going deeper with convolutions.
\newblock In {\em Proceedings of the IEEE conference on computer vision and
  pattern recognition}, pages 1--9, 2015.

\bibitem{tan2019efficientnet}
Mingxing Tan and Quoc Le.
\newblock Efficientnet: Rethinking model scaling for convolutional neural
  networks.
\newblock In {\em International conference on machine learning}, pages
  6105--6114. PMLR, 2019.

\bibitem{tavanaei2020embedded}
Amirhossein Tavanaei.
\newblock Embedded encoder-decoder in convolutional networks towards
  explainable ai.
\newblock {\em arXiv preprint arXiv:2007.06712}, 2020.

\bibitem{touvron2021training}
Hugo Touvron, Matthieu Cord, Matthijs Douze, Francisco Massa, Alexandre
  Sablayrolles, and Herv{\'e} J{\'e}gou.
\newblock Training data-efficient image transformers \& distillation through
  attention.
\newblock In {\em International Conference on Machine Learning}, pages
  10347--10357. PMLR, 2021.

\bibitem{touvron2021going}
Hugo Touvron, Matthieu Cord, Alexandre Sablayrolles, Gabriel Synnaeve, and
  Herv{\'e} J{\'e}gou.
\newblock Going deeper with image transformers.
\newblock In {\em Proceedings of the IEEE/CVF International Conference on
  Computer Vision}, pages 32--42, 2021.

\bibitem{tu2022maxvit}
Zhengzhong Tu, Hossein Talebi, Han Zhang, Feng Yang, Peyman Milanfar, Alan
  Bovik, and Yinxiao Li.
\newblock Maxvit: Multi-axis vision transformer.
\newblock {\em arXiv preprint arXiv:2204.01697}, 2022.

\bibitem{wang2021pyramid}
Wenhai Wang, Enze Xie, Xiang Li, Deng-Ping Fan, Kaitao Song, Ding Liang, Tong
  Lu, Ping Luo, and Ling Shao.
\newblock Pyramid vision transformer: A versatile backbone for dense prediction
  without convolutions.
\newblock In {\em Proceedings of the IEEE/CVF International Conference on
  Computer Vision}, pages 568--578, 2021.

\bibitem{wang2021crossformer}
Wenxiao Wang, Lu Yao, Long Chen, Binbin Lin, Deng Cai, Xiaofei He, and Wei Liu.
\newblock Crossformer: A versatile vision transformer hinging on cross-scale
  attention.
\newblock {\em arXiv preprint arXiv:2108.00154}, 2021.

\bibitem{wang2022semi}
Yuchao Wang, Haochen Wang, Yujun Shen, Jingjing Fei, Wei Li, Guoqiang Jin,
  Liwei Wu, Rui Zhao, and Xinyi Le.
\newblock Semi-supervised semantic segmentation using unreliable pseudo-labels.
\newblock In {\em Proceedings of the IEEE/CVF Conference on Computer Vision and
  Pattern Recognition}, pages 4248--4257, 2022.

\bibitem{wu2021cvt}
Haiping Wu, Bin Xiao, Noel Codella, Mengchen Liu, Xiyang Dai, Lu Yuan, and Lei
  Zhang.
\newblock Cvt: Introducing convolutions to vision transformers.
\newblock In {\em Proceedings of the IEEE/CVF International Conference on
  Computer Vision}, pages 22--31, 2021.

\bibitem{xiao2022delving}
Junfei Xiao, Yutong Bai, Alan Yuille, and Zongwei Zhou.
\newblock Delving into masked autoencoders for multi-label thorax disease
  classification.
\newblock {\em arXiv preprint arXiv:2210.12843}, 2022.

\bibitem{xie2022simmim}
Zhenda Xie, Zheng Zhang, Yue Cao, Yutong Lin, Jianmin Bao, Zhuliang Yao, Qi
  Dai, and Han Hu.
\newblock Simmim: A simple framework for masked image modeling.
\newblock In {\em Proceedings of the IEEE/CVF Conference on Computer Vision and
  Pattern Recognition}, pages 9653--9663, 2022.

\bibitem{yang2022scalablevit}
Rui Yang, Hailong Ma, Jie Wu, Yansong Tang, Xuefeng Xiao, Min Zheng, and Xiu
  Li.
\newblock Scalablevit: Rethinking the context-oriented generalization of vision
  transformer.
\newblock {\em arXiv preprint arXiv:2203.10790}, 2022.

\bibitem{yang2022focal}
Zhendong Yang, Zhe Li, Xiaohu Jiang, Yuan Gong, Zehuan Yuan, Danpei Zhao, and
  Chun Yuan.
\newblock Focal and global knowledge distillation for detectors.
\newblock In {\em Proceedings of the IEEE/CVF Conference on Computer Vision and
  Pattern Recognition}, pages 4643--4652, 2022.

\bibitem{yi2022masked}
Kun Yi, Yixiao Ge, Xiaotong Li, Shusheng Yang, Dian Li, Jianping Wu, Ying Shan,
  and Xiaohu Qie.
\newblock Masked image modeling with denoising contrast.
\newblock {\em arXiv preprint arXiv:2205.09616}, 2022.

\bibitem{yuan2021tokens}
Li Yuan, Yunpeng Chen, Tao Wang, Weihao Yu, Yujun Shi, Zi-Hang Jiang,
  Francis~EH Tay, Jiashi Feng, and Shuicheng Yan.
\newblock Tokens-to-token vit: Training vision transformers from scratch on
  imagenet.
\newblock In {\em Proceedings of the IEEE/CVF International Conference on
  Computer Vision}, pages 558--567, 2021.

\bibitem{zand2022objectbox}
Mohsen Zand, Ali Etemad, and Michael Greenspan.
\newblock Objectbox: From centers to boxes for anchor-free object detection.
\newblock In {\em European Conference on Computer Vision}, pages 390--406.
  Springer, 2022.

\bibitem{zhang2022bending}
Jiaming Zhang, Kailun Yang, Chaoxiang Ma, Simon Rei{\ss}, Kunyu Peng, and
  Rainer Stiefelhagen.
\newblock Bending reality: Distortion-aware transformers for adapting to
  panoramic semantic segmentation.
\newblock In {\em Proceedings of the IEEE/CVF Conference on Computer Vision and
  Pattern Recognition}, pages 16917--16927, 2022.

\bibitem{zhang2022mae}
Kevin Zhang and Zhiqiang Shen.
\newblock i-mae: Are latent representations in masked autoencoders linearly
  separable?
\newblock {\em arXiv preprint arXiv:2210.11470}, 2022.

\bibitem{zhang2022semantic}
Yifan Zhang, Bo Pang, and Cewu Lu.
\newblock Semantic segmentation by early region proxy.
\newblock In {\em Proceedings of the IEEE/CVF Conference on Computer Vision and
  Pattern Recognition}, pages 1258--1268, 2022.

\bibitem{zhao2020covid}
Jinyu Zhao, Yichen Zhang, Xuehai He, and Pengtao Xie.
\newblock Covid-ct-dataset: a ct scan dataset about covid-19.
\newblock {\em arXiv preprint arXiv:2003.13865}, 490, 2020.

\bibitem{zhao2022semantic}
Yizhou Zhao, Xun Guo, and Yan Lu.
\newblock Semantic-aligned fusion transformer for one-shot object detection.
\newblock In {\em Proceedings of the IEEE/CVF Conference on Computer Vision and
  Pattern Recognition}, pages 7601--7611, 2022.

\bibitem{zhou2021deepvit}
Daquan Zhou, Bingyi Kang, Xiaojie Jin, Linjie Yang, Xiaochen Lian, Zihang
  Jiang, Qibin Hou, and Jiashi Feng.
\newblock Deepvit: Towards deeper vision transformer.
\newblock {\em arXiv preprint arXiv:2103.11886}, 2021.

\bibitem{zhou2022rethinking}
Tianfei Zhou, Wenguan Wang, Ender Konukoglu, and Luc Van~Gool.
\newblock Rethinking semantic segmentation: A prototype view.
\newblock In {\em Proceedings of the IEEE/CVF Conference on Computer Vision and
  Pattern Recognition}, pages 2582--2593, 2022.

\bibitem{zhou2022multi}
Wenzhang Zhou, Dawei Du, Libo Zhang, Tiejian Luo, and Yanjun Wu.
\newblock Multi-granularity alignment domain adaptation for object detection.
\newblock In {\em Proceedings of the IEEE/CVF Conference on Computer Vision and
  Pattern Recognition}, pages 9581--9590, 2022.

\end{thebibliography}
}

\clearpage
\appendix

{\LARGE\noindent\textbf{Appendix}}

\vspace{0.5cm}

\section{Datasets}

\begin{table}[ht]
   \centering
   \footnotesize
   \setlength{\tabcolsep}{1.0mm}
   \begin{tabular}{l|c|c|c}   
      Dataset           & Train size        & Test size         & Classes \\
      \hline
    CIFAR-100         & 50000             & 10000             & 100 \\
    CIFAR-10          & 50000             & 10000             & 10 \\
    SVHN              & 73257             & 26032             & 10 \\
    Tiny ImageNet     & 100000            & 10000             & 200 \\
    APTOS 2019        & 3662              & 1928              & 5 \\
    COVID-19          & 546               & 200               & 2 \\
     \end{tabular}
   \caption{The size of datasets used in our experiments.}
\label{tab5}
\end{table} 

In the classification experiment of small datasets, we select Tiny-ImageNet\cite{tavanaei2020embedded}, CIFAR-100\cite{krizhevsky2009learning}, 
CIFAR-10\cite{krizhevsky2009learning} and SVHN\cite{liao2015competitive} as the compared datasets. 
In real-world medical diagnosis application, we choose Aptos 2019\cite{kaggle} and COVID-19\cite{zhao2020covid} as our benchmark datasets. The specific distribution of the datasets is shown in Table~\ref{tab5}.

\vspace{1em}
\noindent CIFAR-100: The CIFAR-100 dataset consists of 32×32 RGB images in real-world. Each class has 600 images, 500 of which are used as training sets and 100 as test sets. 
The 100 classes in CIFAR-100 are divided into 20 superclasses. Each image has a fine label and a coarse label.

\vspace{1em}
\noindent CIFAR-10: The CIFAR-10 dataset consists of 60,000 32×32 color images of 10 classes, with 6000 images for each class. 
It is divided into 50,000 training images and 10,000 test images. Its 10 categories are aircraft, cars, birds, cats, deer, dogs, frogs, horses, boats and trucks.

\vspace{1em}
\noindent SVHN: Street View Door Number (SVHN) dataset is a door number extracted from Google Street View images. Its style is similar to MNIST\cite{lecun1998gradient}, 
and it is also divided into 10 categories. However, it contains a larger order of magnitude of marking data, 
which is used for a more difficult practical problem of recognizing characters and numbers in natural scene images. 
It contains a total of 73257 training images and 26032 test images.

\vspace{1em}
\noindent Tiny-ImageNet: Tiny ImageNet aims to let users solve the problem of image classification as much as possible. 
It contains 100000 images of 200 classes (500 for each class) downsized to 64×64 colored images. 
Each class has 500 training images, 50 validation images and 50 test images.

\begin{figure}[t]
   \centering
   \includegraphics[width=1\linewidth]{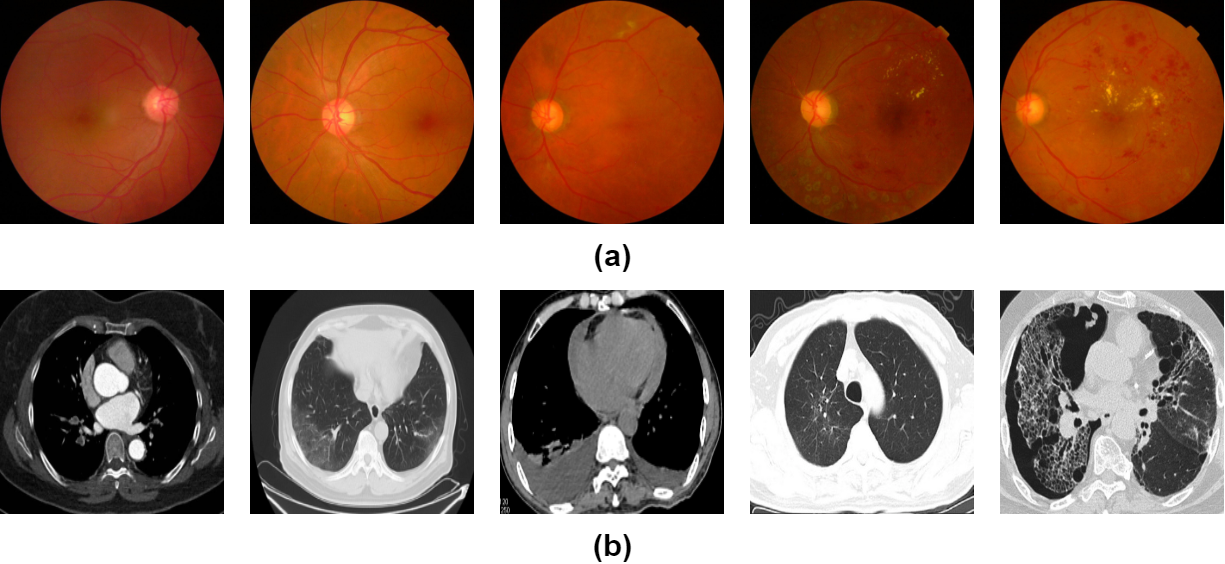}
    \caption{Some examples of our datasets used in medical diagnosis. (a) shows some examples of APTOS 2019. (b) shows some examples of COVID-19.}
    \label{fig11}
 \end{figure}

 \vspace{1em}
 \noindent APTOS 2019: The task of APTOS 2019 is to classify images of diabetes retinopathy into five categories of severity. 
 It uses 3662 high-resolution retinal images as the training set and 1928 images as the test set. Figure~\ref{fig11}(a) shows some examples of APTOS 2019.

 \vspace{1em}
 \noindent COVID-19: It contains the clinical manifestations of COVID-19 from 216 patients. It has a total of 746 CT images, specifically divided into negative and positive. 
 We randomly select 297 negative images and 249 positive images as the training set, 
 and the rest are all test set images. We also show some images of COVID-19 dataset in Figure~\ref{fig11}(b).

 \section{More Results}
 \begin{figure*}[t]
   \centering
   \includegraphics[width=1\linewidth]{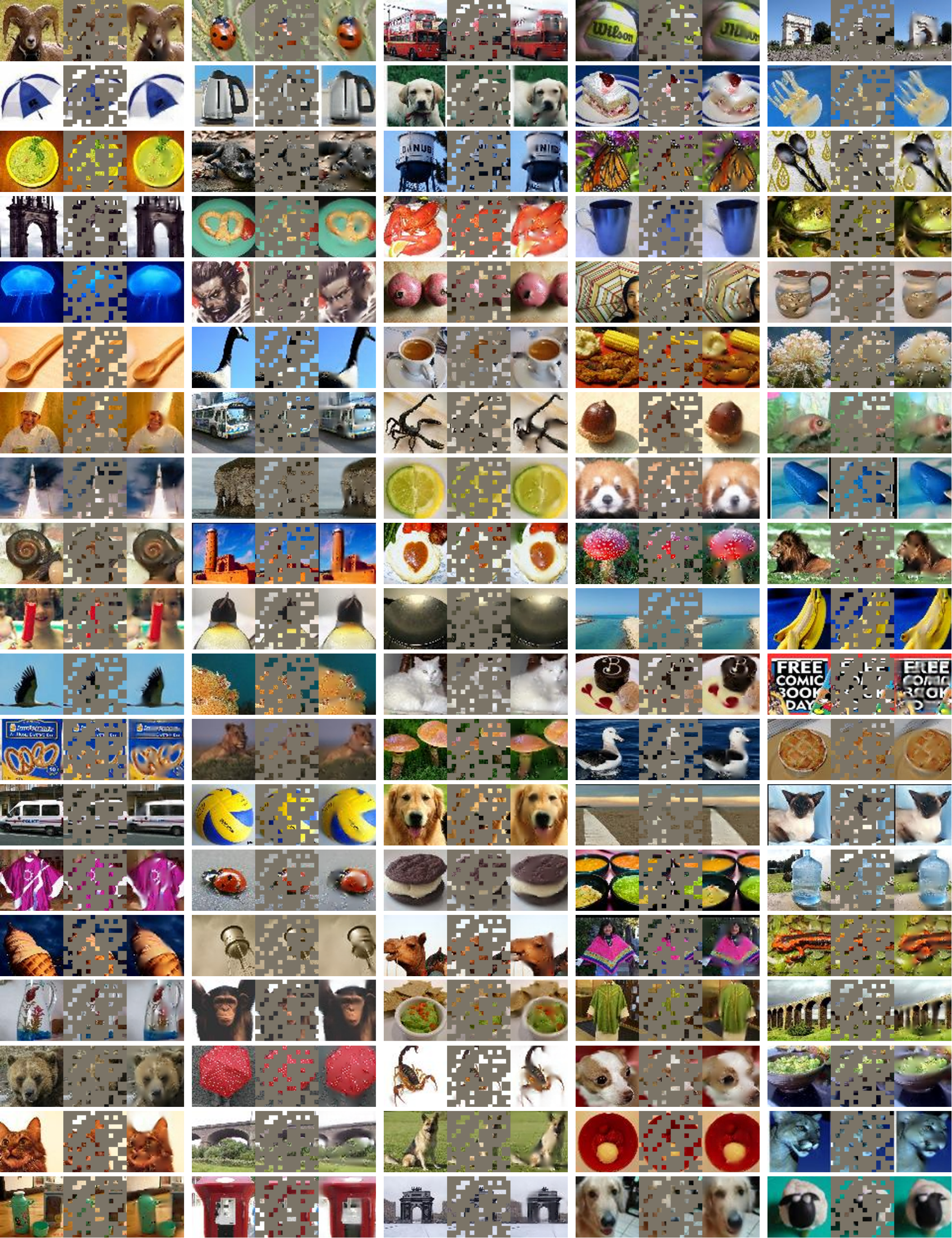}
    \caption{More image prediction results of SDMAE on Tiny-ImageNet. For each triplet, we show the input image (left), the masked image (middle), and prediction result (right). 75\% masking ratio used by us.}
    \label{fig12}
 \end{figure*}

\end{document}